%% file: ms.tex
\documentclass{article}
\pdfoutput=1

\usepackage[utf8]{inputenc} 
\usepackage[T1]{fontenc}    
\usepackage{hyperref}       
\usepackage{url}            
\usepackage{booktabs}       
\usepackage{amsfonts}       
\usepackage{nicefrac}       
\usepackage{microtype}      
\usepackage{lipsum}		
\usepackage{graphicx}
\usepackage{arxiv}

\graphicspath{{pictures/}}
\usepackage{amsmath,amssymb} 
\usepackage{color}
\usepackage[dvipsnames]{xcolor}
\usepackage{xfrac}
\usepackage{comment}
\usepackage{siunitx}
\usepackage{amsmath}
\usepackage{booktabs}
\usepackage{multirow}
\usepackage[ruled,vlined]{algorithm2e}

\usepackage{hyperref}
\usepackage{csquotes}

\title{CAD-Deform: Deformable Fitting of CAD Models to 3D Scans}

\date{} 					

\makeatletter
\newcommand{\printfnsymbol}[1]{%
  \textsuperscript{\@fnsymbol{#1}}%
}
\makeatother

\author{\hspace{1mm}Vladislav Ishimtsev\thanks{Equal contribution} \\
	Skolkovo Institute of Science and Technology\\
	Russia \\
	\And
	Alexey Bokhovkin\printfnsymbol{1} \\
	Skolkovo Institute of Science and Technology\\
	Russia\\
	\And
	Alexey Artemov \\
	Skolkovo Institute of Science and Technology\\
	Russia\\
	\And
	Savva Ignatyev \\
	Skolkovo Institute of Science and Technology\\
	Russia\\
	\And
	Matthias Niessner \\
	Technical University of Munich, Germany\\
	\And
	Denis Zorin \\
	New York University, USA\\
	Skolkovo Institute of Science and Technology\\
	Russia\\
	\And
	Evgeny Burnaev \\
	Skolkovo Institute of Science and Technology\\
	Russia\\
}

\hypersetup{
pdftitle={A template for the arxiv style},
pdfsubject={q-bio.NC, q-bio.QM},
pdfauthor={David S.~Hippocampus, Elias D.~Striatum},
pdfkeywords={First keyword, Second keyword, More},
}

\input tex/99-pre.tex

\begin{document}
\maketitle

\begin{abstract}
Shape retrieval and alignment are a promising avenue towards turning 3D scans into lightweight CAD representations that can be used for content creation such as mobile or AR/VR gaming scenarios. 
Unfortunately, CAD model retrieval is limited by the availability of models in standard 3D shape collections (\eg, ShapeNet). 
In this work, we address this shortcoming by introducing CAD-Deform\footnote[1]{The code for the project: \url{https://github.com/alexeybokhovkin/CAD-Deform}}, a method which obtains more accurate CAD-to-scan fits by non-rigidly deforming retrieved CAD models. 
Our key contribution is a new non-rigid deformation model incorporating smooth transformations and preservation of sharp features, that simultaneously achieves very tight fits from CAD models to the 3D scan and maintains the clean, high-quality surface properties of hand-modeled CAD objects. 
A series of thorough experiments demonstrate that our method achieves significantly tighter scan-to-CAD fits, allowing a more accurate digital replica of the scanned real-world environment while preserving important geometric features present in synthetic CAD environments.
\end{abstract}


\input{tex/1_introduction}
\input{tex/2_related_work}

\input{tex/3_framework_overview}

\input{tex/4_mesh_deformation}
\input{tex/5_datasets}
\input{tex/6_experiments}
\input{tex/9_conclusion}

\clearpage

\bibliographystyle{unsrt}
\bibliography{egbib}

\clearpage

\appendix
\input{tex/9_supplementary}

\clearpage

\end{document}

%% file: tex/99-pre.tex
\def\vs.{vs.\spacefactor=\the\sfcode`\v}
\def\etc.{etc.\spacefactor=\the\sfcode`\c}

\makeatletter
\DeclareRobustCommand\onedot{\futurelet\@let@token\@onedot}
\def\@onedot{\ifx\@let@token.\else.\null\fi\xspace}

\def\eg{\emph{e.g}\onedot} 
\def\ie{\emph{i.e}\onedot} 
\def\wrt{w.r.t\onedot} 
\makeatother

%% file: tex/1_introduction.tex
\section{Introduction}

A wide range of sensors such as the Intel RealSense, Google Tango, or Microsoft Kinect can acquire point cloud data for indoor environments. These data can be subsequently used for reconstructing 3D scenes for augmented and virtual reality, indoor navigation and other applications  \cite{izadi2011kinectfusion,newcombe2011kinectfusion,niessner2013hashing,Whelan15rss,koltun2015reconstruction,dai2017bundle}. 
However, available 3D reconstruction algorithms are not sufficient for many applied scenarios as the quality of the result may be significantly affected by noise, missing data, and other artifacts such as motion blur found in real scans, disabling reconstruction of fine-scale and sharp geometric features of objects. In most instances, reconstructions are still very distant from the clean, 3D models created manually. 

\begin{figure}[ht]
\center{\includegraphics[scale=0.22]{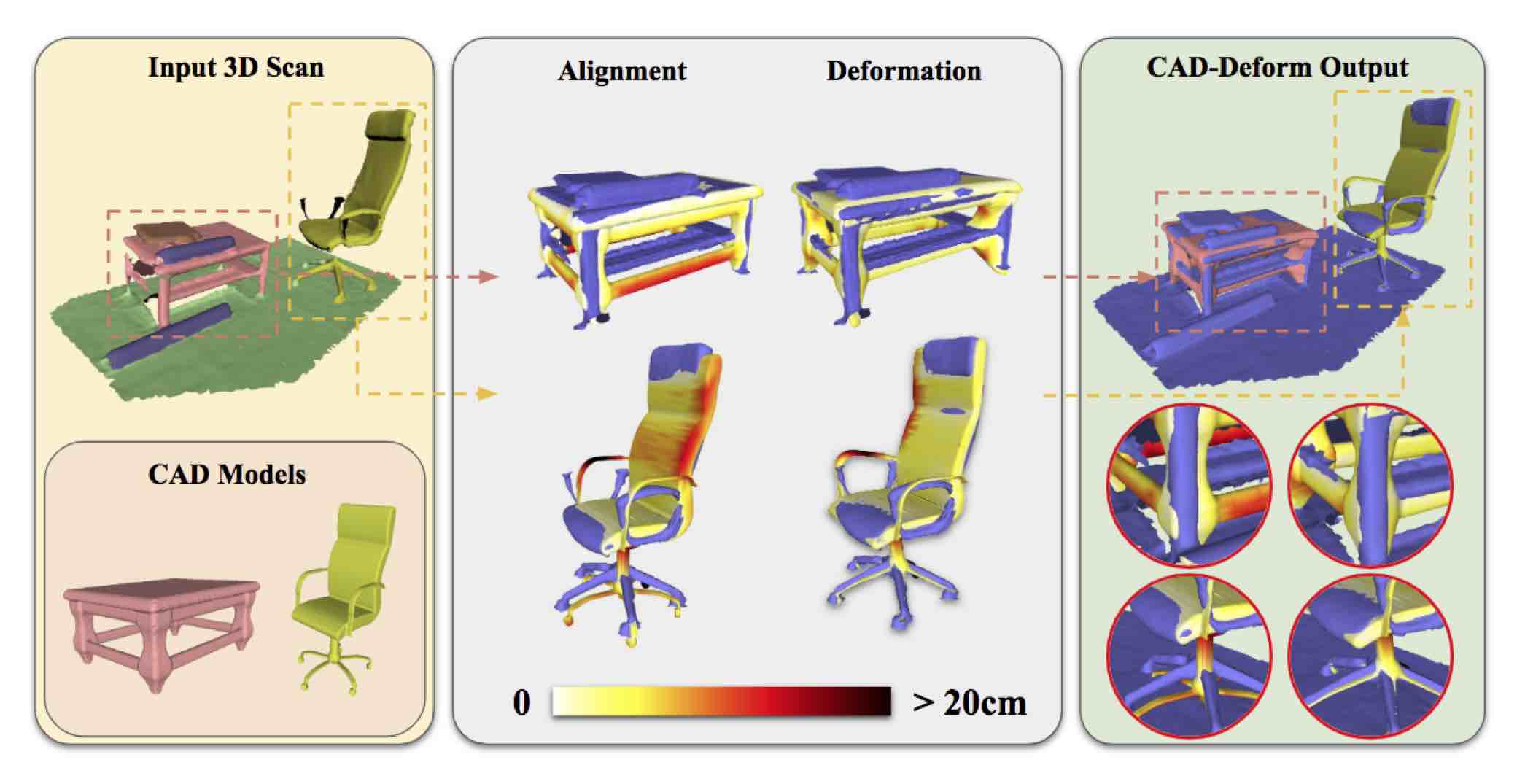}}
\caption{CAD-Deform takes as input a set of 3D CAD models aligned on a RGB-D scan (left). In order to achieve tight fits (middle), we propose a novel part-based deformation formulation that maintains the desired CAD properties such as sharp features.}
\label{fig:teaser}
\end{figure}

An approach to overcome these problems has been proposed in
\cite{moreno2013slam,mattausch2014object} and more recently developed using modern ML methods in \cite{avetisyan2019scan2cad,avetisyan2019end}. Building on the availability of parametric (CAD) models \cite{chang2015shapenet,Koch_2019_CVPR} they perform reconstruction by finding and aligning similar CAD models from a database to each object in a noisy scan. To realize this approach, the authors of \cite{avetisyan2019scan2cad} introduced the Scan2CAD dataset comprising of pairwise keypoint correspondences and 9\,DoF (degrees of freedom) alignments between instances of unique synthetic models from ShapeNet \cite{chang2015shapenet} and reconstructed scans from ScanNet \cite{dai2017scannet}; in order to find and align CAD models to an input scan, they developed a deep neural model to predict correspondences, with a further optimization over potential matching correspondences for each candidate CAD model. The difference of \cite{avetisyan2019end} compared to the approach from \cite{avetisyan2019scan2cad} is an end-to-end procedure, combining initially decoupled steps to take into account additional feedback through the pipeline by learning the correspondences specifically tailored for the final alignment task. 

However, geometric fidelity achieved between scans and CAD objects remains limited. CAD model geometry (clean and complete) differs significantly from scan geometry in low-level geometric features. As these methods focus on finding alignments by optimizing a small number of parameters (9\,DoF), resulting alignments only roughly approximate scans, not capturing geometric details such as variation in 3D shapes of parts of individual objects. 

In contrast, to improve geometric fidelity while keeping the benefit of mesh-based representations, we propose to increase the number of degrees of freedom by allowing the CAD objects to \emph{deform} rather than stay rigid. 
In this work, we introduce a deformation framework CAD-Deform, which significantly increases the geometric quality of object alignment regardless of the alignment scheme. For an input scan, given object location and 9\,DoF alignment of a potentially matching CAD model, we apply a specially designed mesh deformation procedure resulting in a more accurate shape representation of the underlying object.
The deformation matches each semantic part of a 3D shape extracted from the PartNet~\cite{mo2019partnet} to the corresponding data in the scans and keeps sufficient rigidity and smoothness to produce perceptually acceptable results while minimizing the distance to scanned points. Thus, even if the initial location and alignment are not entirely accurate, the deformation can compensate to a significant extent for the discrepancy.

Our approach builds highly detailed scene descriptions with a high level of semantic accuracy for applications in 3D graphics.
The approach outperforms state-of-the-art methods for CAD model alignment and mesh deformation by 2.1--6.3\% for real-world 3D scans.  To the best of our knowledge, the approach we propose is the first to use mesh deformation for scan-to-CAD alignment and real-world scene reconstruction. In summary, our work has several contributions:
\begin{itemize}
\item We developed a mesh deformation approach that 1) is computationally efficient, 2) does not require exact correspondences between each candidate CAD model and input scan, and 3) provides perceptually plausible deformation thanks to a specially introduced smoothness term and inclusion of geometric features of a CAD model in the optimization functional. 

\item We developed a methodology to assess the fitting accuracy and the perceptual quality of the scan-to-CAD reconstruction. The methodology includes standard data fitting criteria similar to Chamfer distance to evaluate alignment accuracy, complimentary local and global criteria for visual quality assessment of resulting deformations, and a user study.

\item We performed an ablation study to assess the influence of inaccuracies in the initial object location and alignment on the final reconstruction results. For that we used both ground-truth alignments from Scan2CAD dataset~\cite{avetisyan2019scan2cad} along with predictions of their method, and alignments trained in the end-to-end fashion~\cite{avetisyan2019end}. We compared results with the state-of-the-art methods for mesh deformation to highlight the advantages of our approach.
\end{itemize}

%% file: tex/2_related_work.tex
\section{Related work}
\label{sec:related_work}

\textit{RGB-D scanning and reconstruction.}  RGB-D scanning and reconstruction are increasingly widely used, due to the availability of commodity range sensors and can be done both in real-time and offline modes. There are many methods for RGB-D-based real-time reconstruction such as KinectFusion~\cite{izadi2011kinectfusion}, Voxel Hashing~\cite{niessner2013hashing} or Bundle Fusion~\cite{dai2017bundle} that use the well-known volumetric fusion approach from~\cite{volumetricfusion}. ElasticFusion~\cite{Whelan15rss} is a representative offline approach to the reconstruction. These methods can produce remarkable results for large 3D environments. However, due to occlusions and imperfections of existing sensors, the reconstructed scenes contain many artifacts such as noise, missing surface parts, or over-smooth surfaces. Some methods aim to predict unobserved or corrupted parts of 3D scans from depth data. In~\cite{firman2016structured}, the world is modeled as a grid of voxels representing the signed distance to the nearest surface. Then structured Random Forest is used to predict the value of the signed distance function for each of the voxels computed to form a final occupancy estimate for unobserved voxels. \cite{song2017semantic} is to encode a depth map as a 3D volume and then aggregate both local geometric and contextual information via a 3D CNN to produce the probability distribution of voxel occupancy and object categories for all voxels inside the camera view frustum. Another approach is~\cite{dai2017shape}, where a 3D CNN architecture predicts a distance field from a partially-scanned input and finds the closest 3D CAD model from a shape database. By copying voxels from the nearest shape neighbors, they construct a high-resolution output from the low-resolution predictions, hierarchically synthesize a higher-resolution voxel output, and extracts the mesh from the implicit distance field. However, although all these methods can complete partial scans and improve 3D geometry reconstruction, the quality of the results still is far from artist-created 3D content.

\textit{CAD model alignment.} Instead of reconstructing 3D geometry in a bottom-up manner, one can perform reconstruction by retrieving CAD models from a dataset and aligning them to the noisy scans. Matching CAD models to scans requires extracting 3D feature descriptors; thus, approaches have been proposed for 3D feature extraction. Hand-crafted features are often based on various histograms of local characteristics (\eg,~\cite{drost2012ppf}). Such approaches do not generalize well to inherent variability and artifacts in real-world scans. Deep learning approaches lead to further improvements: \eg,~\cite{zhou2018viewconsistency} propose a view consistency loss for a 3D keypoint prediction network based on RGB-D data; \cite{deng2019local} develop 3D local learnable features for pairwise registration. After the descriptors are extracted, one can use a standard pipeline for CAD-to-scan alignment: first, matches between points based on 3D descriptors are identified, and then variational alignment is used to compute 6- or 9-DoF CAD model alignments. Typical examples, realizing this two-step approach, are described in~\cite{moreno2013slam,mattausch2014object,yangyan2015database,avetisyan2019scan2cad,avetisyan2019end}. The most recent ones~\cite{avetisyan2019scan2cad,avetisyan2019end} use learnable descriptors and differ in that the latter considers an end-to-end scan-to-CAD alignment, reconstructing a scene in a single forward pass. An obvious limitation of these two-step pipelines is that the resulting alignments approximate scans only coarsely due to a pre-defined set of available CAD models and a highly constrained range of transformations. Other approaches in this category~\cite{Aubry14,guo2015predicting,gupta2015aligning}, although relying on the same two-step strategy,  use only a single RGB or RGB-D input.

\textit{Mesh deformation.} To improve surface reconstruction accuracy and obtain a more faithful description of geometric details (\eg, preserve distinctive geometric features), it is desirable to consider a more general space of deformations than a simple non-uniform scaling. 
To this end, a mesh deformation approach based on Laplacian surface editing is presented in~\cite{stoll2006template}. 
Another iterative mesh deformation scheme~\cite{sorkine2007rigid} imposes rigidity on the local transformations. 
Despite their conceptual simplicity, both methods require specifying correspondences between mesh vertices and scan data points. The same is true for~\cite{liao2009deform,achenbach2015accurate,amberg2007icp} and unsuitable in our setting, due to extremely low (2--8) number of correspondences available per part, that additionally may not even be well defined for noisy and incomplete real-world scans.
Many methods~\cite{achenbach2015accurate,cagniart2009iterative,park2014template,dey2015automatic} focus on automatic posing of human models, dense surface tracking and similar applications. 
However, while producing compelling results, the methods implicitly use the assumption that a template mesh and its target shape are very similar; as a result, the semantic parts of individual 3D objects are not changed either in shape or relative scale. 
In our work, we are comparing to ARAP (as-rigid-as-possible)~\cite{sorkine2007rigid} and Harmonic~\cite{botsch2004intuitive,jacobson2010mixed} mesh deformation methods, however, with an added Laplacian smoothing term to leverage second-order information about the surface. This modification makes ARAP/Harmonic similar to~\cite{achenbach2015accurate,liao2009deform} as far as non-data-dependent energy terms are concerned. 
Other methods exist that propose non-linear constraints~\cite{he2013l0,grinspun2003shells,frohlich2011shells}. 
The energy terms of our framework were designed as a natural match for our problem: we define local 3D transformations on the CAD model, mapping each subpart to the 3D scene volume, and require smooth changes of these transformations over the model. In contrast to most deformation methods, we do not aim to keep the surface close to isometric: \eg, a table reference model can be stretched highly anisotropically to match a different table in the data. Our energy is defined using local 3D affine transform primitives, penalizing sharp changes, without penalizing anisotropic deformations. These primitives also allow us to express 1D feature preservation simply, and the non-data terms are quadratic which is critical for our efficient preconditioned optimization. Methods in~\cite{he2013l0,grinspun2003shells,frohlich2011shells} propose non-linear energies, focusing on large rotations, but implicitly assuming quasi-isometry; adapting these methods is nontrivial.

%% file: tex/3_framework_overview.tex
\section{Overview of CAD-Deform framework}
\label{sec:hloverview}

Our approach is built on top of the framework from~\cite{avetisyan2019scan2cad,avetisyan2019end} for CAD model retrieval and 9\,DoF alignment in 3D scans.
By running any of the approaches from \cite{avetisyan2019scan2cad,avetisyan2019end} for an input scan, we obtain initial object locations and 9\,DoF alignments of CAD models potentially matching specific parts of the scan. 
Next, we apply our proposed mesh deformation procedure (see Section \ref{sec:method}), resulting in a more accurate shape representation of the aligned objects: 
\begin{enumerate}

    \item We segment the CAD models into semantic 3D parts following the labelling from the PartNet dataset~\cite{mo2019partnet}.

    \item For each aligned object, we select points in the scan that are the nearest (within some fixed radius) to each vertex of the CAD model. 
    We assign a label of the nearest part of the aligned CAD model to each such point.

    \item As an input to the proposed mesh deformation procedure, we use the mesh model with semantic part labels and labelled segment of the 3D scene.

    \item We deform the mesh by optimizing the energy depending on the relative positions of mesh vertices and labelled points of the scene, see Section~\ref{sec:method}.

\end{enumerate}

%% file: tex/4_mesh_deformation.tex
\begingroup

\newcommand{\bM}{\mathbf{M}}
\newcommand{\bE}{\mathbf{E}}
\newcommand{\bV}{\mathbf{V}}
\newcommand{\bF}{\mathbf{F}}
\newcommand{\bP}{\mathbf{P}}
\newcommand{\bQ}{\mathbf{Q}}
\newcommand{\bC}{\mathbf{C}}
\newcommand{\bG}{\mathbf{G}}
\newcommand{\bW}{\mathbf{W}}
\newcommand{\bL}{\mathbf{L}}
\newcommand{\cE}{\cal{E}}
\newcommand{\R}{\mathbb{R}}

\section{Data-driven shape deformation}
\label{sec:method}

\begin{figure}[!t]
\center{\includegraphics[scale=0.12]{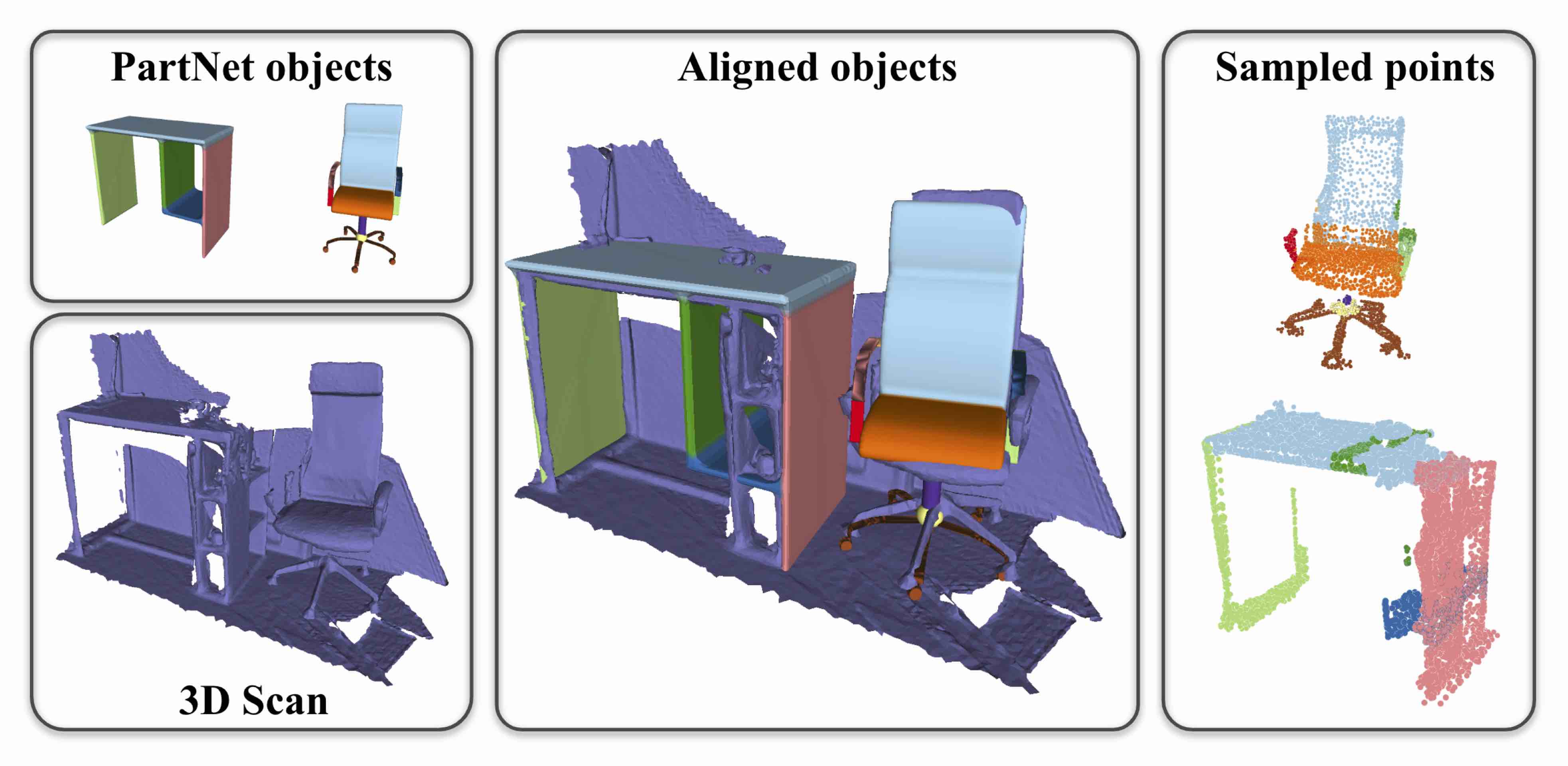}}
\caption{Data acquisition for CAD-Deform: we project PartNet labels onto aligned ShapeNet CAD models (left), register these models to a 3D scene scan (middle), and extract points on the scene within $\varepsilon$-neighborhood of aligned mesh surface (we set $\varepsilon = 10$\,cm), copying labels corresponding to nearest part of CAD model (right).}
\label{fig:input_example}
\end{figure}

In this section, we describe our method for fitting a closest-match mesh from the shape dataset to the scanned data. Our algorithm assumes two inputs: 
\begin{itemize}
    \item an initial mesh $\bM$ from the dataset, with part labels assigned to vertices and 9~degrees-of-freedom (9\,DoF) transformation assigned to mesh;
    \item a subset of the scanned data, segmented with the same part labels as $\bM$. 
\end{itemize}
 Fig.~\ref{fig:input_example} shows an example of the input data, see also Section~\ref{sec:hloverview}.
  
\textit{Notations.} A mesh  $\bM = (\bV^0, \bE, \bF)$ consists of a set of initial 3D vertex coordinates $\bV^0$ of size $n_v$, the edge set $\bE$ of size $n_e$, and the triangle face set $\bF$ of size $n_f$.   We compute the deformed vertex positions $\bV$ by minimizing a deformation energy. 
All vertices are assigned with part labels, which is a map  $\bQ: \bV^0 \to \bC$, where $\bC$ is the set of labels $c_i, i=1, \ldots, n_c$, and $n_c$ is the number of parts in all objects in our dataset.
For a mesh $\bM_m$, $\bC_{m} \subset \bC$ is the set of labels of its parts.  The set of  $n_{\bP}$ points  $\bP$ has the same labels as the mesh $\bM_m$ we fit, \ie,  $\bC_{\bP} = \bC_m$. 
In addition, we assume that for every mesh $\bM_m$ on the scan we have a scaling transformation, represented by a $4 \times 4$ matrix $T^0_m$, that aligns it with voxelized points $\bP$.
In our optimization,  we use \emph{per-edge} transformations $T_e$, 
discussed below, which we use to measure the deviation from a scaled version of the original shape and deformation smoothness.

\input{tex/4.1_mesh_deformation_energy}

\input{tex/4.2_mesh_deformation_quadratic}

\input{tex/4.3_mesh_deformation_dataterm}

\input{tex/4.4_mesh_deformation_opt}

\endgroup

%% file: tex/4.1_mesh_deformation_energy.tex
\subsection{Deformation energy}
\label{method:energy}

Our goal is to define a deformation energy to match the scanned data as closely as possible, while maintaining continuity and deformation smoothness, and penalizing deviation from the original shape.   In addition, we include a term that preserves perceptually important linear geometric features of the mesh.

Conceptually, we follow common mesh deformation methods such as ARAP~\cite{sorkine2007rigid} which estimate a local transformation from vertex positions and penalize the deviation of this transformation from the target (\eg, closest rotation). An important distinction is that the transformation we need to estimate locally is a 3D, rather than 2D transformation, and cannot be estimated from the deformed positions of vertices of a single triangle.

\textit{Edge transformations.}
We associate local 3D affine transformations with mesh edges.
Each transformation is defined in a standard way by a $4 \times4 $ matrix in projective coordinates, with the last row $(0,0,0,c)$, $c \neq 0$.
The vertices of two faces $(i_1,i_2,i_3)$ and $(i_2,i_1,i_4)$, incident at an edge $e = (i_1,i_2)$, form
a (possibly degenerate) tetrahedron.
If it is not degenerate, then there is a unique linear transformation $T_e$ mapping the undeformed positions $(v^0_{i_1}, v^0_{i_2}, v^0_{i_3}, v^0_{i_4})$ to  $(v_{i_1}, v_{i_2}, v_{i_3}, v_{i_4})$.
Matrix $T^0$ of the affine transformation has 12 coefficients that are uniquely determined by the equations $T_e v^0_i = v_i$, $i=1\ldots 4$; moreover, these are linear functions of the deformed positions $v_i$, as these are only present on the right-hand side of the equations.
Handling of degenerate tetrahedra is discussed in Section~\ref{method:smooth-sharp}.

\textit{Energy.} We define the following non-linear objective for the unknown deformed vertex positions $\bV$ and (a fixed) point set $\bP$:
\begin{align}
\cE(\bV, \bP)
    &= \underbrace{
        \boxed{
            E_{\text{shape}}
            + \alpha_{\text{smooth}} E_{\text{smooth}}
            + \alpha_{\text{sharp}} E_{\text{sharp}}
        }
    }_{\text{quadratic problem}}
            + \alpha_{\text{data}} E_{\text{data}}, \notag  \\
E_{\text{shape}}
    &= \underbrace{
        \sum_{e \in \bE} \|T_e(\bV) -T_e^0\|^2_2
    }_{\text{deviation}};
\qquad
E_{\text{smooth}} =
    \sum_{f \in \bF} \sum_{e_i, e_j \in f} \|T_{e_i}(\bV) - T_{e_j}(\bV)\|^2_2; \notag \\
E_{\text{sharp}} &=
    \sum_{k = 1}^{n_p} \sum_{e_s \in \bE_{\text{sharp}}^k}\|T_{e_s}(\bV) - T_{e_{s+1}}(\bV)\|^2_2;
\qquad
E_{\text{data}} =
    f_{\text{data}}(\bV, \bP).\label{eq:main_energy}
\end{align}

The first term penalizes deviations of the 3D affine transformations defined by the deformed vertex positions from the transformation (a non-uniform scale) that aligns mesh with the data. This term directly attempts to preserve the shape of the object, modulo rescaling.

As explained above, the transformations $T_e(\bV)$ are defined for non-degenerate input mesh configurations. Suppose the four initial vertex positions $(v^0_{i_1}, v^0_{i_2}, v^0_{i_3}, v^0_{i_4})$ form a degenerate tetrahedron $\bW$, \ie, two faces incident at the edge are close to co-planar. In this case, we use an energy term consisting of two terms defined per triangle. Instead of $4 \times 4$ matrix for each non-degenerate tetrahedron, there is a $3 \times 2$ matrix for each triangular face of degenerate tetrahedron that represents a transformation restricted to the plane of this face.
Note that in this case, the deformation in the direction perpendicular to the common plane of the triangles does not affect the energy, as it does not have an impact on the local shape of the deformed surface. We explain the remaining energy terms in the next sections.

%% file: tex/4.2_mesh_deformation_quadratic.tex
\subsection{Quadratic terms}
\label{method:smooth-sharp}

\textit{Smoothness term.} This term can be thought of as a discrete edge-based Laplacian, applied to the transforms associated with edges: the difference of any pair of transforms for edges belonging to the same triangle is penalized. 

\textit{Sharp features term.}  We have observed that a simple way to preserve some of the perceptually critical geometric aspects of the input meshes is to penalize the change of deformations $T_e$ along sharp geometric features, effectively preserving their overall shape (although still allowing, possibly non-uniform, scaling, rotations and translations). We detect sharp edges based on a simple dihedral angle thresholding, as this is adequate for the classes of CAD models we use.
Detected sharp edges are concatenated in sequences, each consisting of vertices with exactly two incident sharp edges each, except the first and the last. Those sequences are defined for each part of the mesh.  Sequences of different parts has no common sharp edges or vertices belonging to them. Effectively, the sharpness term increases the weights for some edges in the smoothness term.

%% file: tex/4.3_mesh_deformation_dataterm.tex
\begingroup

\newcommand{\onetoone}{one-to-one~}
\newcommand{\Onetoone}{One-to-one~}
\newcommand{\oto}{o2o}
\newcommand{\Nneighbor}{Nearest-neighbor~}
\newcommand{\nneighbor}{nearest-neighbor~}
\newcommand{\nn}{nn}
\newcommand{\idx}{\iota}

\newcommand{\prox}{d}  
\newcommand{\scr}{\xi} 
\newcommand{\chr}{H} 

\subsection{Data term}
\label{method:dataterm}

We use two approaches to defining the data term, one based on screened attraction between close points in the mesh $\bM_m$ and $\bP$, and the other one based on attraction between \emph{a priori} chosen corresponding points of the mesh $\bM_m$ and data points in the set $\bP$.
We found that the former method works better globally, when the deformed mesh is still far from the target point cloud, while the latter is able to achieve a better match once a closer approximation is obtained.

\textit{Part-to-part mapping.} We define a data-fitting term based on point proximity: we set an energy proportional to the distance between sufficiently close points. To avoid clustering of mesh points, we add a screening term that disables attraction for mesh vertices close to a given point. 

We denote by $H(x)$ the Heaviside function, \ie, a function with value $1$ if $x \geqslant 0$ 
and zero otherwise. We set
\begin{equation}
\label{eq:data_term_p2p}
f_{\text{data}}^{\text{p2p}}(\bV, \bP) = 
	\sum_{c \in \bC} \sum_{v \in \bV_c} \sum_{p \in \bP_c} 
		\scr^{\sigma}(p, \bV_c)\big(\prox^{\varepsilon}(v - p)\big)^2,  
\end{equation}
where $\prox^{\varepsilon}(v - p)  = (v - p) \cdot H(\varepsilon- \|v - p\|^2_2)$,
and $\scr^{\sigma}(p, \bV) = H(\min_{\{v \in \bV\}} \|v - p\| - \sigma)$ is a \enquote{screening} function.
The value for $\sigma$ is chosen to be the mean edge length, the value of $\varepsilon$ is about $10$ edge lengths. To make $f_{\text{data}}^{\text{p2p}}$ differentiable, instead of the Heaviside function and $\min$, we can use their smoothed approximations. 

\textit{\Nneighbor mapping.} Recall that the vertices \{$v_i$\} of the mesh $m$ and the points \{$p_i$\} of the set $\bP$ have the same part label sets $\bC_m = \bC_{\bP}$. 
For each label $c$, we consider $\bV^0_c$ and $\bP_c$, the set of mesh vertices in initial positions and the set of points with the same label $c$ in $\bP$. 
Let $B_{\bV_c}, B_{\bP_c}$ be the bounding boxes of these sets, and consider the affine transform $T_{B}^c$ that maps $B_{\bV_c}$ to $B_{\bP_c}$.
Among all possible correspondences between corners of the boxes, we choose the one that produces the affine transform closest to identity. 
Then the index $i = \idx^T(p)$ of the vertex $v_i$, corresponding to a point $p \in \bP_c$, is determined as $\idx^c(p) = \arg\min_{\{i, v_i \in \bV_c\}} \|T_B^c v_i - p\|^2_2.$
Then  the data term is defined as
\begin{equation}
\label{eq:data_term_nn}
f_{\text{data}}^{\text{nn}}(\bV, \bP) = 
	\sum_{c \in \bC} \sum_{p \in \bP_c}
		\|p - v_{\idx^{c}(p)}\|^2_2. 
\end{equation}

\endgroup

%% file: tex/4.4_mesh_deformation_opt.tex
\begingroup

\newcommand{\vp}{\overline{\bV}}

\subsection{Optimization}
\label{method:optimization}

The optimization of~\eqref{eq:main_energy} is highly nonlinear due to the data term.
However, all other terms are quadratic functions of vertex coordinates, so minimizing 
\begin{align*}
E_{\text{quad}} 
&= 	E_{\text{shape}}
	+ \alpha_{\text{smooth}} E_{\text{smooth}}
	+ \alpha_{\text{sharp}} E_{\text{sharp}} \\ 
&=  \vp^{\top} A_{\text{shape}} \vp
    + \alpha_{\text{smooth}} \vp^{\top} A_{\text{smooth}} \vp
    + \alpha_{\text{sharp}} \vp^{\top} A_{\text{sharp}} \vp + b^{\top} \vp
\end{align*}
is equivalent to solving a system of linear equations.  
Denoting the sum of the matrices in the equation by $A_{\text{quad}}$, we obtain the optimum by solving
$A_{\text{quad}} \vp = 0$ where the vector $\vp$ is a flattening of the vector $\bV$ of 3D vertex positions to a vector of length $3n_{\bV}$. 

The data term is highly nonlinear, but solving the complete optimization problem can be done efficiently using $A_{\text{quad}}^{-1}$ as the preconditioner.
For our problem, we use the preconditioned L-BFGS optimizer summarized in Algorithm~\ref{alg:bfgs}.  
\begin{algorithm}[H]
\SetAlgoLined
\DontPrintSemicolon
{
$M_{\text{precond}} = A_{\text{quad}}^{-1}$  \tcp{stored as LU decomposition} 
$\vp = T_{m}^0(\vp^0)$ \\
 \For{$i \gets 0$ \KwTo $N_{\text{iter}}$}{
    $g_{\text{tot}} = 
   		\alpha_{\text{data}} \sfrac{\mathrm{d} E_{\text{data}}}{\mathrm{d}p}
      + A_{\text{quad}} \vp + b$\;
    $\vp = \text{L-BFGS-step}(\vp, g_{\text{tot}}, M_{\text{precond}})$
    }
}
\caption{Preconditioned L-BFGS mesh optimization (PL-BFGS)}
\label{alg:bfgs}
\end{algorithm}

%% file: tex/5_datasets.tex
\section{Datasets}
\label{sec:datasets}

Our method relies on a number of publicly available datasets to compute mesh annotations for our deformations and assess fitting performance. To assess fitting performance, we use Scan2CAD dataset~\cite{avetisyan2019scan2cad} that consists of 14225~ground-truth 9\,DoF transformations between objects in 1506~reconstructions of indoor scenes from Scannet~\cite{dai2017scannet} and 3049~unique CAD models from ShapeNet~\cite{chang2015shapenet}. 

Our deformation framework requires high-quality watertight meshes to support numerically stable optimization, which does not hold for ShapeNet CAD models. Thus, we remesh these to around 10k--15k vertices using~\cite{huang2018robust}, obtaining more uniform discretizations. To annotate these remeshed CAD models with semantic part labels required by our deformation procedure, we register them with the corresponding part-labeled meshes from the PartNet dataset~\cite{mo2019partnet} by first choosing an appropriate \ang{90} rotation around each axis and then optimizing for a finer alignment using a point-to-point ICP algorithm~\cite{rusinkiewicz2001efficient} between vertices of the two meshes. We annotate the semantic parts for vertices in each mesh by projecting it from the respective closest vertices of the registered PartNet mesh.

The original ShapeNet meshes, however, are used to extract sharp geometric features, as these have easily detectable sharp angles between adjacent faces. We label as sharp all edges adjacent to faces with a dihedral angle smaller than the threshold $\alpha_{\text{sharp}} = \ang{120}$. We further project vertex-wise sharpness labels from the original to the remeshed CAD models and select a sequence of edges forming the shortest paths between each pair of vertices as sharp.

%% file: tex/6_experiments.tex
\begingroup

\section{Results}
\label{sec:experiments}

\input{tex/6.1_experiments_setup}

\input{tex/6.2_experiments_fitting}

\input{tex/6.3_experiments_cad_quality}

\input{tex/6.4_experiments_ablation}

\input{tex/6.5_experiments_morphing}

\endgroup

%% file: tex/6.1_experiments_setup.tex
\subsection{Evaluation setup}
\label{experiments:setup}

Our performance evaluation of obtained deformations is multifaceted and addresses the following quality-related questions: 
\begin{itemize}
    \item Scan fitting performance: \emph{How well do CAD deformations fit?}
    
    \item Perceptual performance for deformations: \emph{How CAD-like are deformations?}
    
    \item Contributions of individual energy terms: \emph{Which energy terms are essential?}
    
    \item Deformation flexibility: \emph{Can better shapes be achieved by approximating clean meshes rather than noisy scans?}
\end{itemize}

\textit{Fitting and perceptual metrics.} We quantify the deformation performance in terms of fitting quality between the scene scans and 3D CAD models using a family of related measures computed on a per-instance basis. For vertices $\bV = (v_i)$ of the deformed mesh $\bM$, we compute distances to their respective nearest neighbors $(\text{NN}(v_i, \mathbf{S}))$ in the scan $\mathbf{S}$. We compute per-instance $\text{Accuracy} = |\bV_{\text{close}}| / |\bV|$, reflecting the fraction of closely located vertices, and trimmed minimum matching distance $\text{tMMD} = \sum\limits_{v_i \in \bV} \min(\tau, \|v_i - \text{NN}(v_i, \mathbf{S})\|_1) / |\bV|$, where  $\bV_{\text{close}} = \{v_i \in \bV: \|v_i - \text{NN}(v_i, \mathbf{S})\|_1 < \tau\}$ is the set of vertices falling within $L_1$-distance $\tau$  to their nearest neighbor in the scan, and $\tau$ controls robustness \wrt incomplete scans. We set $\tau = 0.2$ (see supplementary) in our experiments and report Accuracy and tMMD values averaged over classes and instances. 

There is no universally agreed perceptual quality measure for meshes; thus, we opted for a tripartite evaluation for our resulting deformations. First, we measure dihedral angle mesh error (DAME)~\cite{dame2012}, revealing differences in \textit{local surface quality} between the original and the distorted meshes:
\[\text{DAME}(\mathbf{M}, \mathbf{M}_{\text{def}}) = \frac{1}{|\mathbf{E}|} 
    \sum\limits_{\text{adjacent}f_1, f_2}
    \big|D_{f_1, f_2} - \overline{D_{f_1, f_2}}\big| \cdot 
        \exp\big\{(Z_{\text{DAME}} D_{f_1, f_2})^2\big\},
\]
where $D_{f_1, f_2}$ and $\overline{D_{f_1, f_2}}$ represent oriented dihedral angles between faces $f_1$ and $f_2$ in the original and deformed meshes, respectively, and $Z_{\text{DAME}} = \sfrac{\sqrt{\log(100 / \pi)}} {\pi}$ is a parameter scaling DAME values to $[0, 100]$.

Second, we assess \textit{abnormality of deformed shapes} with respect to the distribution of the undeformed shapes, building on the idea of employing deep autoencoders for anomaly detection in structured high-dimensional data. By replicating the training instances, autoencoders learn features that minimize reconstruction error; for novel instances \emph{similar} to those in the training set, reconstruction error is low compared to that of strong \emph{outliers}. We train six autoencoders~\cite{achlioptas2018learning,lapgan2019} for point clouds using vertices of undeformed meshes separately for the top six classes present in Scan2CAD annotation: \textit{table, chair, display, trashbin, cabinet,} and \textit{bookshelf}. 
Passing vertices $\bV_{\text{def}}$ of a deformed shape to the respective autoencoder, one can assess how accurately deformed meshes can be approximated using features of undeformed meshes. This property can be evaluated with Earth Mover's Distance (EMD) $d_{\text{EMD}}(\bV_{\text{def}}, \bV'_{\text{def}}) = \min\limits_{\phi: \mathbf{\bV_{\text{def}}} \to \mathbf{\bV'_{\text{def}}}} \sum\limits_{v \in \bV_{\text{def}}} \| v - \phi(v) \|_2,$
where $\phi$ is a bijection, obtained as a solution to the optimal transportation problem involving $\bV_{\text{def}}$ and $\bV'_{\text{def}}$, that can intuitively be viewed as the least amount of work needed to transport $\bV_{\text{def}}$ vertices to positions of $\bV'_{\text{def}}$.

Lastly, we assess \textit{real human perception} of deformations in a user study, detailed in Section~\ref{exper:cad_quality}.

\begin{figure}[t!]
\center{\includegraphics[scale=0.29]{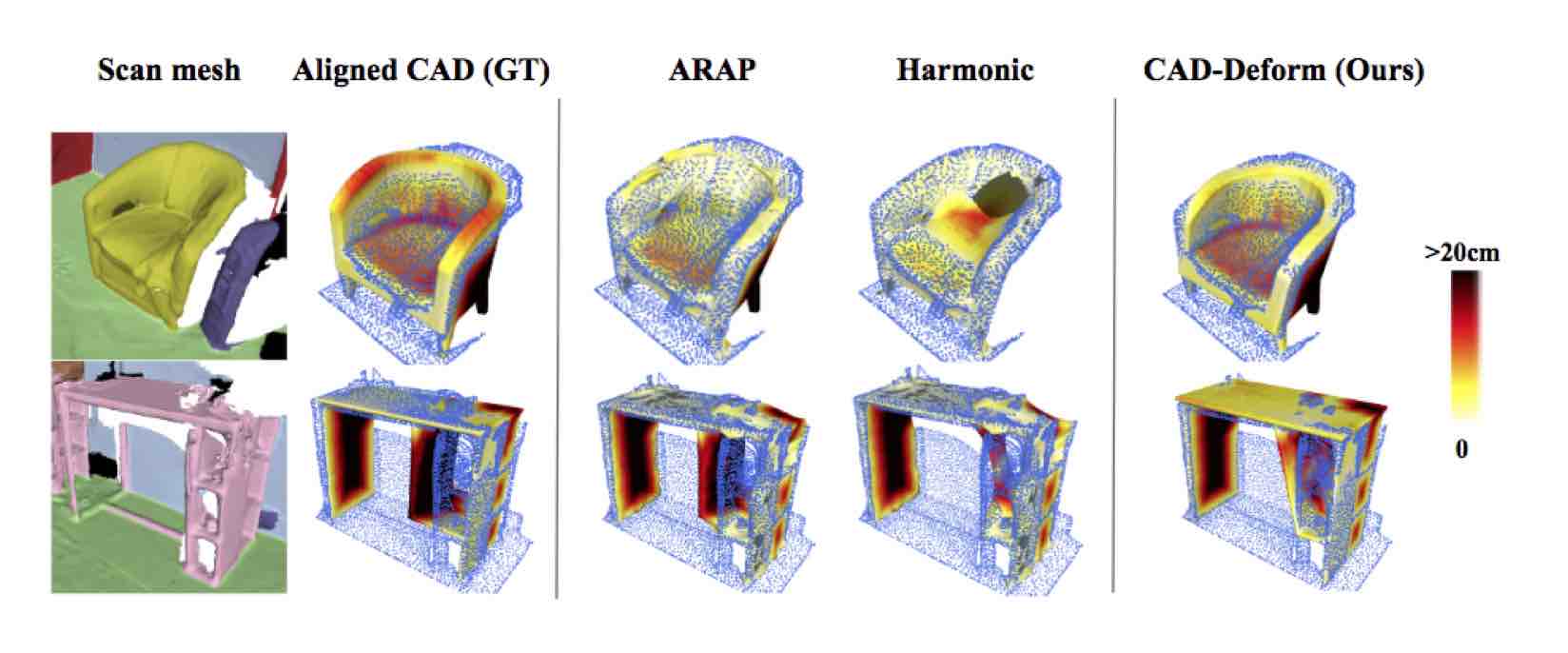}}
\caption{Deformations obtained using our method and the baselines, with mesh colored according to the Euclidean distance to its nearest point in the scan. We note that high accuracy scores for Harmonic and ARAP deformations are achieved at the cost of destroying the initial structure of the mesh, particularly in regions where scan is missing (note that back side and armrests are gone for chairs in the first and second rows). In contrast, our method is better able to preserve smooth surfaces, sharp features, and overall mesh integrity, while keeping accurate local alignment.}
\label{fig:fitting_performance}
\end{figure}

\textit{Optimization details.} To perform quantitative comparisons, we use 299~scenes in ScanNet constituting Scan2CAD validation set~\cite{avetisyan2019scan2cad}, but with 697~shapes present in PartNet dataset, amounting to 1410 object instances. Our full experimental pipeline is a sequence of deformation stages with different optimization parameters, and Hessian being recomputed before each stage. Specifically, we perform one \textit{part-to-part} optimization with parameters $\alpha_{\text{shape}} = 1, \alpha_{\text{smooth}} = 0, \alpha_{\text{sharp}} = 0, \alpha_{\text{data}} = 5\times 10^{4}$ for 100~iterations, then we perform $5$~runs of \textit{nearest-neighbor} deformation for 50~iterations with parameters $\alpha_{\text{shape}} = 1, \alpha_{\text{smooth}} = 10, \alpha_{\text{sharp}} = 10, \alpha_{\text{data}} = 10^3$. More details about optimization and timings are provided in the supplementary. 

%% file: tex/6.2_experiments_fitting.tex
\subsection{Fitting Accuracy: How well do CAD Deformations fit?}
\label{exper:fitting-perf}

\begin{table}[t]
\centering
\resizebox{0.8\textwidth}{!}{%
\begin{tabular}{l ccc ccc}
\toprule
    & \multicolumn{3}{c}{\textbf{Class avg.}} 
    & \multicolumn{3}{c}{\textbf{Instance avg.}} \\
    \textbf{Method}
    & \textbf{GT} 
    & \textbf{S2C}~\cite{avetisyan2019scan2cad} 
    & \textbf{E2E}~\cite{avetisyan2019end} 
    & \textbf{GT} 
    & \textbf{S2C}~\cite{avetisyan2019scan2cad} 
    & \textbf{E2E}~\cite{avetisyan2019end} \\
\midrule
\# TPs          & 1410 & 499 & 882 & 1410 & 499 & 882 \\
TP undeformed    & 89.2 & 83.7 & 88.5 & 90.6 & 79.4 & 93.9 \\
\midrule
Ours: NN only       & 89.7 & 84.3 & 89.0 & 91.4 & 84.7 & 94.4 \\
Ours: p2p only      & 90.3 & 88.3 & 89.4 & 91.6 & 90.3 & 94.9 \\
Ours: w/o smooth & 90.6 & \textbf{90.0} & 89.6 & 92.3 & 90.3 & 95.0  \\
Ours: w/o sharp     & 90.3 & 86.9 & \textbf{90.6} & 92.3 & 89.4 & \textbf{95.2} \\
CAD-Deform           & \textbf{91.7} & 89.8 & 90.3 & \textbf{93.1} & \textbf{92.8} & 94.6 \\
\bottomrule
\end{tabular}
}
\caption{Comparative evaluation of our deformations to true positive (TP) alignments by non-deformable approaches in terms of Accuracy (\%). Note that deformations improve performance for all considered alignment approaches.}
\label{tab:deform_comparison}

\centering
\resizebox{1.0\textwidth}{!}{%
\begin{tabular}{l c c c c c c c c c}
\toprule
\textbf{Method} & bookshelf & cabinet & chair & display & table & trashbin & other & ~~class avg. & avg. \\
\# instances & 142 & 162 & 322 & 86 & 332 & 169 & 197 & 201.4 & 1410 \\ \midrule
Ground-truth                   & 88.0 & 75.2 & 94.8 & 98.9 & 89.6 & 96.6 & 81.4 & 89.2 & 90.6 \\
Ours                           & \textbf{90.5}  & \textbf{82.2} & \textbf{95.4} & \textbf{99.1} & \textbf{91.0} & \textbf{98.6} & \textbf{84.8} & \textbf{91.7} & \textbf{93.1} \\ \bottomrule 
\end{tabular}
}
\caption{Comparative evaluation of our approach to non-deformable ground-truth and baselines in terms of scan approximation Accuracy (\%). We conclude that our deformations improve fitting accuracy across all object classes by~2.5\,\% on average.}
\label{tab:fitting_performance}
\end{table}

We first demonstrate how deformation affects scan fitting performance for meshes aligned using different methods, specifically, we use true-positive shape alignments computed using Scan2CAD (S2C)~\cite{avetisyan2019scan2cad}, End-to-End (E2E)~\cite{avetisyan2019scan2cad}, as well as ground-truth alignments. We start with an aligned mesh, copy the 9\,DoF transformation to each of the mesh vertices, and optimize using our deformation method with parameters described in Section~\ref{experiments:setup}. We report Accuracy scores in terms of fraction of well-approximated points in the scan for aligned shapes pre- and post-optimization in Table~\ref{tab:fitting_performance}, achieving improved performance across all considered alignment procedures.

Surprisingly, we improve even over ground-truth alignments by as much as 2.5\,\%. Thus, we compute per-class scores in Table~\ref{tab:deform_comparison} (comparison across alignments), reporting improvements of up to 7\,\% in average scan approximation accuracy that are consistent \textit{across all object classes}.
We visualize example deformations obtained using our approach and baselines in Figure~\ref{fig:fitting_performance}.

We have discovered our deformation framework to be robust \wrt level of detail in the data term in~\eqref{eq:data_term_p2p} and provide more detail in the supplementary.

%% file: tex/6.3_experiments_cad_quality.tex
\subsection{CAD Quality: How CAD-like are deformed models?}
\label{exper:cad_quality}

Having obtained a collection of deformed meshes, we aim to assess their visual quality in comparison to two baseline deformation methods: as-rigid-as-possible (ARAP)~\cite{sorkine2007rigid} and Harmonic deformation~\cite{botsch2004intuitive,jacobson2010mixed}, using a set of perceptual quality measures. To bring second-order information about mesh surface in energy formulation of ARAP/Harmonic, we add the Laplacian smoothness term $E_{\text{Lap}}(\bV, \bV') = \sum_{i=1}^{|\bV|}\|L(v_i)-L(v_i')\|^2,$ where $L(v_i) = \frac{1}{\mathcal{N}_i}\sum_{j=1}^{|\mathcal{N}(i)|}v_j$ and $\mathcal{N}(i)$ is a set of one-ring neighbors of vertex $v_i\in\bV$ ($v_i'\in\bV'$). The details of our user study design and visual assessment are provided in the supplementary.

\bgroup
\setlength\tabcolsep{0.5em}
\begin{table}[b]
\centering
\resizebox{0.8\textwidth}{!}{%
\begin{tabular}{l cc cc c cc}
\toprule
    & \multicolumn{2}{c}{\textbf{DAME}}
    & \multicolumn{2}{c}{\textbf{EMD} $\times 10^{-3}$}
    & \textbf{User}
     & \multicolumn{2}{c}{\textbf{Accuracy}} \\
    \textbf{Method}
    & cls.
    & inst.
    & cls.
    & inst.
    & \textbf{study}
    & cls.
    & inst. \\
\midrule
No deformation   & 0 & 0 & 77 & 77 & 8.6 & 89.2 & 90.6 \\
\midrule
ARAP \cite{sorkine2007rigid}   & 47.1 & 45.7 & 88 & 87 & 4.0 & 90.8 & 91.8 \\
Harmonic \cite{botsch2004intuitive,jacobson2010mixed}        & 65.1 & 65.2 & 104 & 102 & 2.6 & \textbf{96.2} & \textbf{96.6} \\
CAD-Deform       & \textbf{20.5} & \textbf{17.2} & \textbf{84} & \textbf{84} & \textbf{7.7} & 91.7 & 93.1 \\
\bottomrule
\end{tabular}
}
\caption{Quantitative evaluation of visual quality of deformations obtained using ARAP~\cite{sorkine2007rigid}, Harmonic deformation~\cite{botsch2004intuitive,jacobson2010mixed}, and our CAD-Deform, using a variety of local surface-based (DAME~\cite{dame2012}), neural (EMD~\cite{achlioptas2018learning,lapgan2019}), and human measures.}
\label{tab:big_picture}
\end{table}
\egroup


%% file: tex/6.4_experiments_ablation.tex
\subsection{Ablation study}
\label{exper:ablative}

\begin{figure}[ht!]
\center{\includegraphics[scale=0.24]{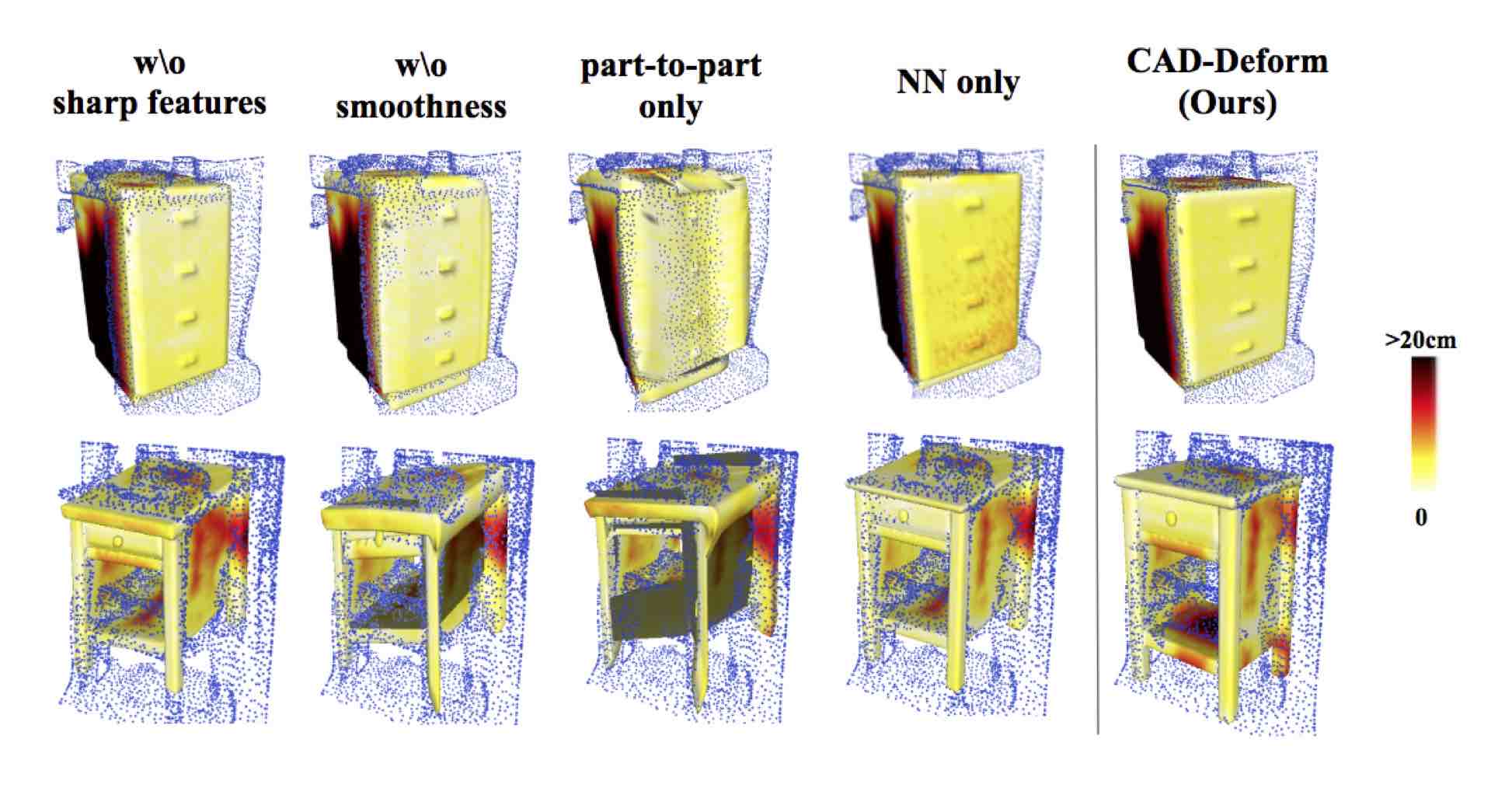}}
\caption{Qualitative results of ablation study usind our deformation framework, with mesh coloured according to the value of the tMMD measure. Note that including smoothness term is crucial to prevent surface self-intersections, while keeping sharpness allows to ensure consistency in parallel planes and edges.}
\label{fig:ablation_result}
\end{figure}

To evaluate the impact of individual energy terms in~\eqref{eq:main_energy} on both scan fitting performance and perceptual quality, we exclude each term from the energy and compute deformations by optimizing the remaining ones. First, we exclude sharpness or smoothness terms, optimizing for deformations using the original two-stage method; second, to better understand the influence of each stage, we perform experiments with only the first or the second stage (a single run only). We aggregate results into Table~\ref{tab:deform_comparison} and display them visually in Figure~\ref{fig:ablation_result}, concluding that our CAD-Deform maintains the right balance between fit to the scan and perceptual quality of resulting deformations. 

\begin{figure}[ht!]
\center{\includegraphics[scale=0.26]{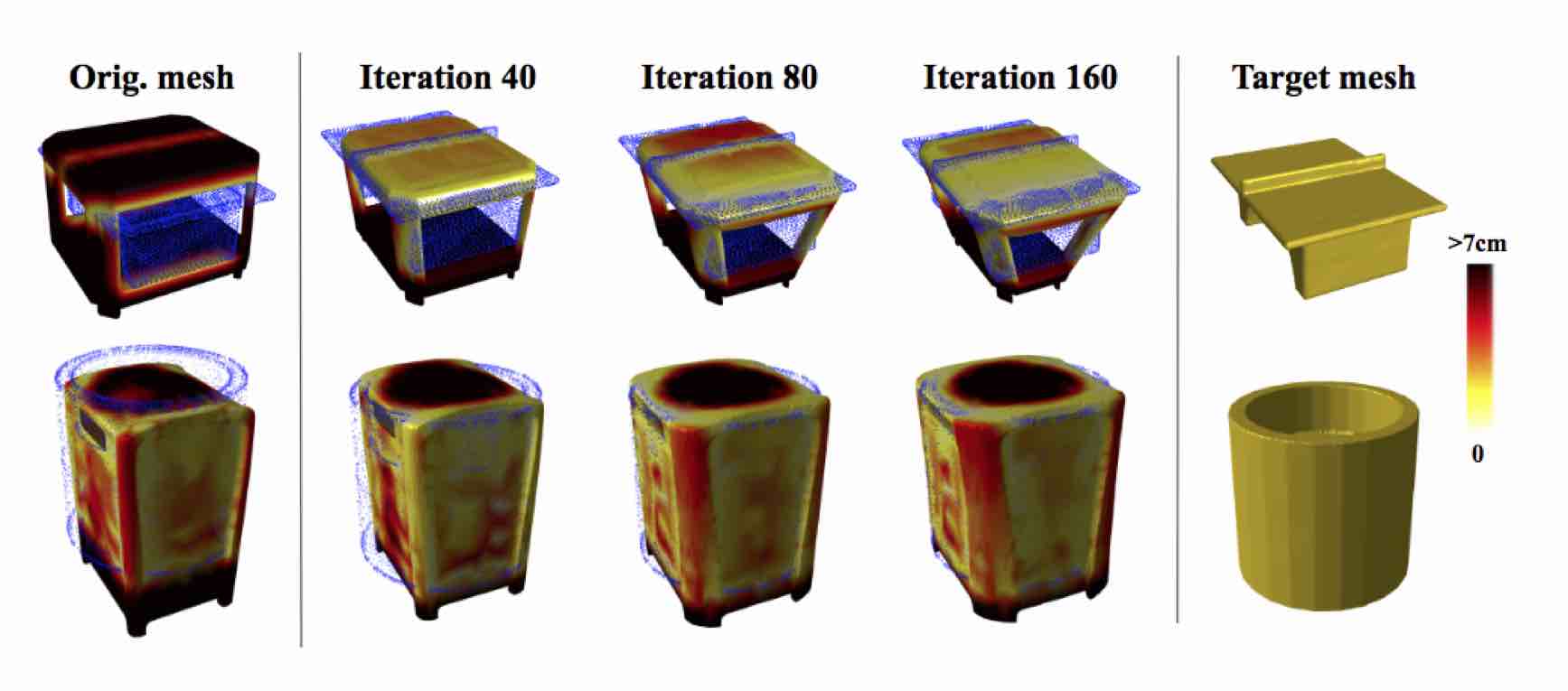}}
\caption{Qualitative shape translation results, interpolating between the original mesh (left) and the target mesh (right). }
\label{fig:morphing_example}
\end{figure}

%% file: tex/6.5_experiments_morphing.tex
\subsection{Shape morphing results}
\label{exper:morphing}

To demonstrate the ability of our mesh deformation framework to perform shape interpolation, we choose two different meshes in the same ShapeNet category and optimize our energy~\eqref{eq:main_energy} to approximate one with the other, see Fig.~\ref{fig:morphing_example}.

%% file: tex/9_conclusion.tex
\section{Conclusion}
\label{sec:conclution}

In this work, we have presented CAD-Deform, a shape deformation-based scene reconstruction method leveraging CAD collections that is capable of improving over existing alignment methods.
More specifically, we introduce a composite deformation energy formulation that achieves regularization from semantic part structures, enforces smooth transformations, and preserves sharp geometric features.
As a result we obtain significantly improved perceptual quality of final 3D CAD models compared to state-of-the-art deformation formulations, such as ARAP and polyharmonic deformation frameworks.
Overall, we believe that our method is an important step towards obtaining lightweight digital replica from the real world that are both of high-quality and accurate fits at the same time.


\section*{Acknowledgements} 
\label{sec:acknowledgements}

The authors acknowledge the usage of the Skoltech CDISE HPC cluster Zhores for obtaining the results presented in this paper. The work was partially supported by the Russian Science Foundation under Grant 19-41-04109.

%% file: tex/9_supplementary.tex

\input{tex/suppl_A_dataset_intersection}
\input{tex/suppl_B_optimization_details}
\input{tex/suppl_C_fitting}

\input{tex/suppl_G_morphing}
\input{tex/suppl_H_parts}
\input{tex/suppl_I_threshold_optimization}
\input{tex/suppl_J_user_study}

%% file: tex/suppl_A_dataset_intersection.tex
\section{Statistics on the used datasets}
\label{supsec:dataset_stats}

In Tables~\ref{tab:object_presence}\,\&\,\ref{tab:class_presence}, we summarize statistical information on the number of instances and categories considered in our evaluation. As we require parts annotations as an important ingredient in our deformation, we only select instances in Scan2CAD~\cite{avetisyan2019scan2cad} where the associated parts annotation in PartNet~\cite{mo2019partnet} is available, resulting in total in 9~categories (25\%), 572~instances (18\%), and 1979~annotated correspondences (14\%). Note that the vast majority of cases remain within our consideration, keeping our evaluation comprehensive.



\begin{table}[ht!]
\centering
\resizebox{0.8\textwidth}{!}{%
\begin{tabular}{l cccc}
\toprule
\textbf{Collection} & \textbf{Categories} & \textbf{Instances} & \textbf{Corresp.} \\
\midrule
Scan2CAD~\cite{avetisyan2019scan2cad} & 35 & 3,049 & 14,225 \\
           \qquad w/parts annotations & 24 & 2,477 & 12,246 \\
\bottomrule
\end{tabular}
}
\caption{Overall statistics on the numbers of categories, instances, and correspondences present in our datasets.}
\label{tab:object_presence}
\end{table}

\begin{table}[ht!]
\centering
\resizebox{0.75\textwidth}{!}{%
\begin{tabular}{l cc cc}
\toprule
    \multirow{2}{*}{\textbf{Name}}
    & \multicolumn{2}{c}{\textbf{Scan2CAD}}
    & \multicolumn{2}{c}{\textbf{PartNet $\cap$ Scan2CAD}}  \\
    & corresp.
    & shapes
    & corresp. 
    & shapes \\
\midrule
\multicolumn{5}{l}{\textit{Shape categories used in our evaluation:}} \\
chair & 4677 & 652 & 4351 & 632 \\
table & 2616 & 830 & 2594 & 822 \\
cabinet & 1401 & 310 & 1258 & 294 \\
trash bin & 1044 & 89 & 1042 & 88 \\
bookshelf & 824 & 150 & 812 & 145 \\
display & 770 & 165 & 762 & 161 \\
\midrule
\multicolumn{5}{l}{\textit{Shape categories NOT used in our evaluation:}} \\
bed & 355 & 50 & 342 & 47 \\
file cabinet & 294 & 70 & 290 & 68 \\
bag & 165 & 9 & 165 & 9 \\
lamp & 135 & 55 & 135 & 55 \\
bathtub & 474 & 96 & 129 & 25 \\
microwave & 99 & 37 & 98 & 36 \\
sofa & 577 & 247 & 60 & 20 \\
laptop & 51 & 24 & 51 & 24 \\
keyboard & 62 & 11 & 48 & 9 \\
\bottomrule
\end{tabular}
}
\caption{The top~15~most frequent ShapeNet categories in Scan2CAD dataset including a detailed information on those with the availability of the corresponding parts annotations.}
\label{tab:class_presence}
\end{table}

We further select the most well-presented six shape categories as our core evaluation set, outlined in Table~\ref{tab:class_presence}. Note that as our method is non-learnable, we can just as easily experiment with the remaining categories, at the cost of somewhat reduced statistical power.


%% file: tex/suppl_B_optimization_details.tex
\section{Optimization details}
\label{supsec:optimization}

Our full experimental pipeline is a sequence of deformation stages with different optimization parameters, and Hessian being recomputed before each stage. Specifically, we perform one \textit{part-to-part} optimization with parameters $\alpha_{\text{shape}} = 1, \alpha_{\text{smooth}} = 0, \alpha_{\text{sharp}} = 0, \alpha_{\text{data}} = 5\times 10^{4}$ for 100~iterations, then we perform $5$~runs of \textit{nearest-neighbor} deformation for 50~iterations with parameters $\alpha_{\text{shape}} = 1, \alpha_{\text{smooth}} = 10, \alpha_{\text{sharp}} = 10, \alpha_{\text{data}} = 10^3$. 
Such number of iterations was sufficient to achieve convergence with energy changes less than $10^{-1}$ in our experiments. Runtime of our method breaks into cost computation ($\mathtt{\sim}0.3$\,s), backward ($\mathtt{\sim}0.2$\,s), and optimization steps containing the main bottleneck (sparse matrix-vector multiplication) ($\mathtt{\sim}1.2$\,s) for a typical $10^4$\,vertices mesh. All operations can be easily further optimized.

%% file: tex/suppl_C_fitting.tex
\section{Qualitative fitting results}
\label{supsec:more_shape_fitting}

\begin{figure}[ht]
\center{\includegraphics[scale=0.26]{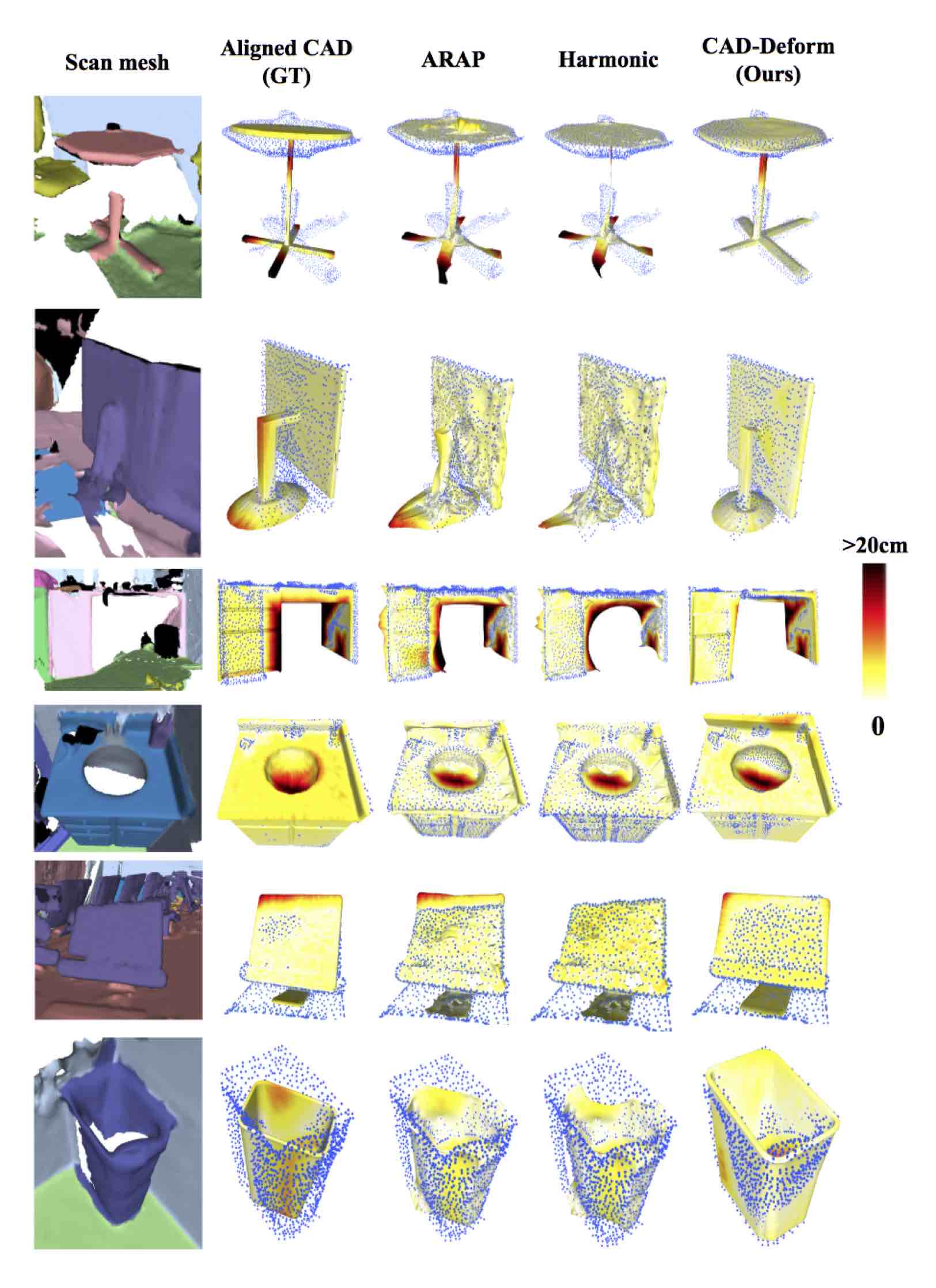}}
\caption{Qualitative shape deformation results using obtained using ARAP~\cite{sorkine2007rigid}, Harmonic deformation~\cite{botsch2004intuitive,jacobson2010mixed}, and our CAD-Deform. Mesh surface is colored according to the value of tMMD measure, with darker values corresponding to the larger distance values.}
\label{fig:fitting_examples}
\end{figure}

\begin{table}[ht!]
\centering
\resizebox{1.0\textwidth}{!}{%
\begin{tabular}{l c c c c c c c c c }
\toprule
\textbf{Method} & {bookshelf} & {cabinet} & {chair} & {display} & {table} & {trash bin} & {other} & ~~{class avg.} & {avg.} \\
\midrule
ARAP      & 52.48 & 41.77 & 45.52 & 51.30 & 41.77 & 57.00 & 39.75 & 47.08 & 45.67 \\
Harmonic  & 64.77 & 58.74 & 68.06 & 64.22 & 58.26 & 80.13 & 61.70 & 65.12 & 65.18 \\
\midrule
Ours: NN only     & \textbf{21.54} & \textbf{23.39} & \textbf{7.31} & \textbf{18.37} & \textbf{18.69} & 18.13 & \textbf{16.07} & \textbf{17.64} & \textbf{14.14} \\
Ours: p2p only    & 22.44 & 24.28 & 9.51 & 21.12 & 18.76 & \textbf{15.30} & 18.34 & 18.54 & 15.57  \\
Ours: w/o smooth & 27.15 & 29.27 & 14.50 & 27.05 & 24.48 & 24.39 & 23.26 & 24.30 & 20.95 \\
Ours: w/o sharp & 26.43 & 25.98 & 13.34 & 24.87 & 22.47 & 21.04 & 21.18 & 22.19 & 19.10 \\
CAD-Deform & 24.8 & 24.1 & 11.4 & 24.4 & 21.6 & 19.4 & 17.6 & 20.5 & 17.2 \\
\bottomrule 
\end{tabular}
}
\caption{Quantitative results of local surface quality evaluation using DAME measure~\cite{dame2012} (the smaller, the better, normalized to a maximum score of 100), where our CAD-Deform compares favourably to the baselines across all considered categories. Note, however, how surface quality significantly decreases when smoothness and sharp feature-related terms are dropped.}
\label{tab:dame_performance}
\end{table}

\begin{table}[ht!]
\centering
\resizebox{0.9\textwidth}{!}{%
\begin{tabular}{l cccccc cc}
\toprule
\textbf{EMD} $\times 10^{-3}$ & bookshelf & cabinet & chair & display & table & trash bin & class avg. & avg. \\
\midrule
Ground-truth & 77.8 & 78.9 & 76.1 & 77.5 & 77.3 & 73.1 & 76.8 & 77.0 \\
\midrule
ARAP \cite{sorkine2007rigid} & 80.3 & 86.5 & 88.5 & \textbf{85.4} & \textbf{86.8} & 98.1 & 87.6 & 87.3 \\
Harmonic \cite{botsch2004intuitive,jacobson2010mixed} & 94.0 & 110.6 & 95.8 & 95.3 & 103.2 & 122.6 & 103.6 & 101.7 \\
CAD-Deform & \textbf{79.0} & \textbf{81.1} & \textbf{80.3} & 91.7 & 87.0 & \textbf{87.4} & \textbf{84.4} & \textbf{83.8} \\
\bottomrule 
\end{tabular}
}
\caption{Results of LSLP-GAN reconstruction in terms of Earth-Mover's Distance between reconstructed and original point clouds of mesh vertices.}
\label{tab:reconstruction_performance}
\end{table}

\begin{figure}[ht!]
\center{\includegraphics[scale=0.18]{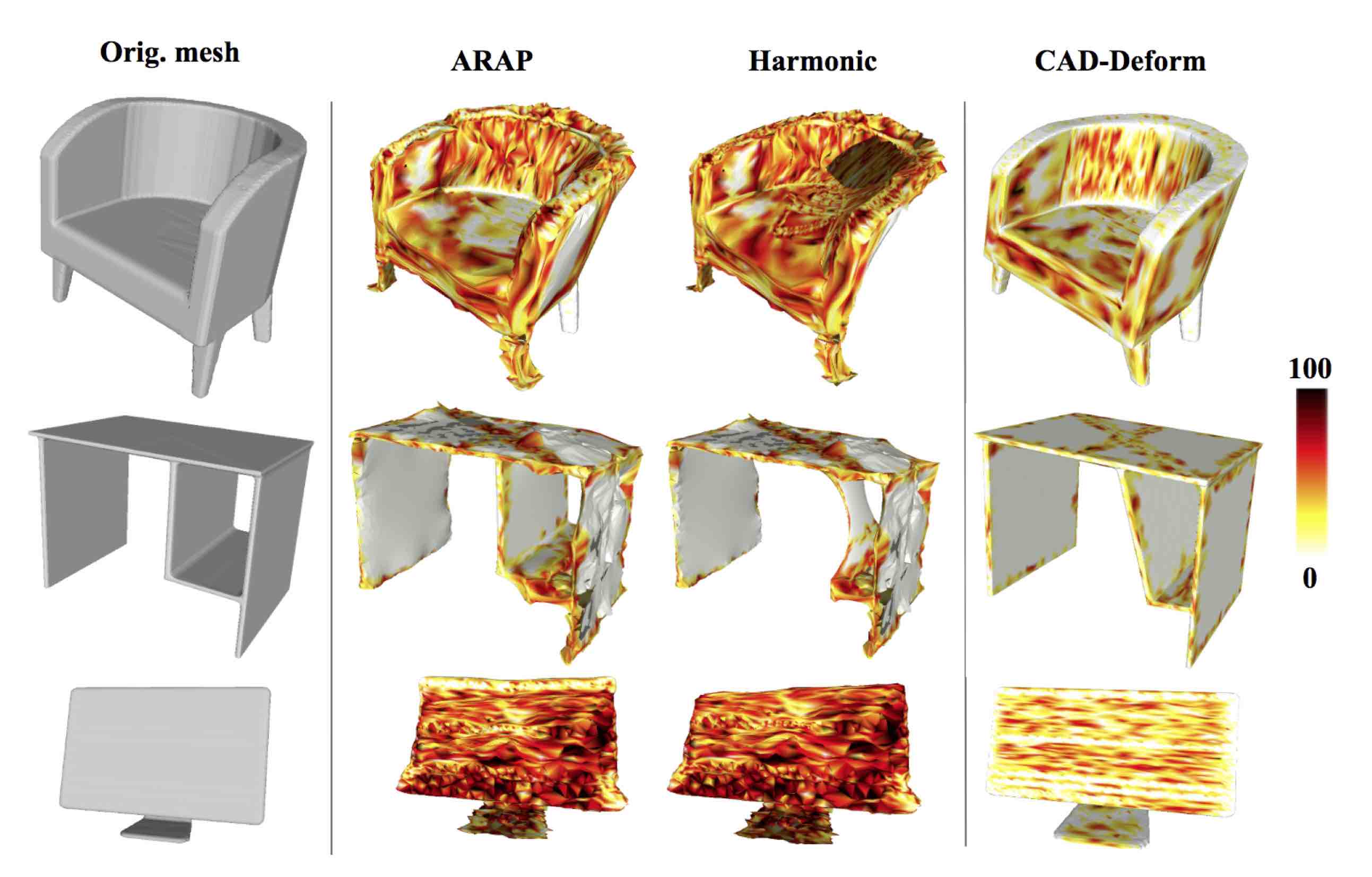}}
\caption{Qualitative comparison of deformations obtained using ARAP~\cite{sorkine2007rigid}, Harmonic deformation~\cite{botsch2004intuitive,jacobson2010mixed}, and our CAD-Deform, with shapes coloured according to the value of DAME measure~\cite{dame2012}. Our approach results in drastic improvements in local surface quality, producing higher-quality surfaces compared to other deformations.}
\label{fig:dame_performance}
\end{figure}

\begin{figure}[ht]
\center{\includegraphics[scale=0.26]{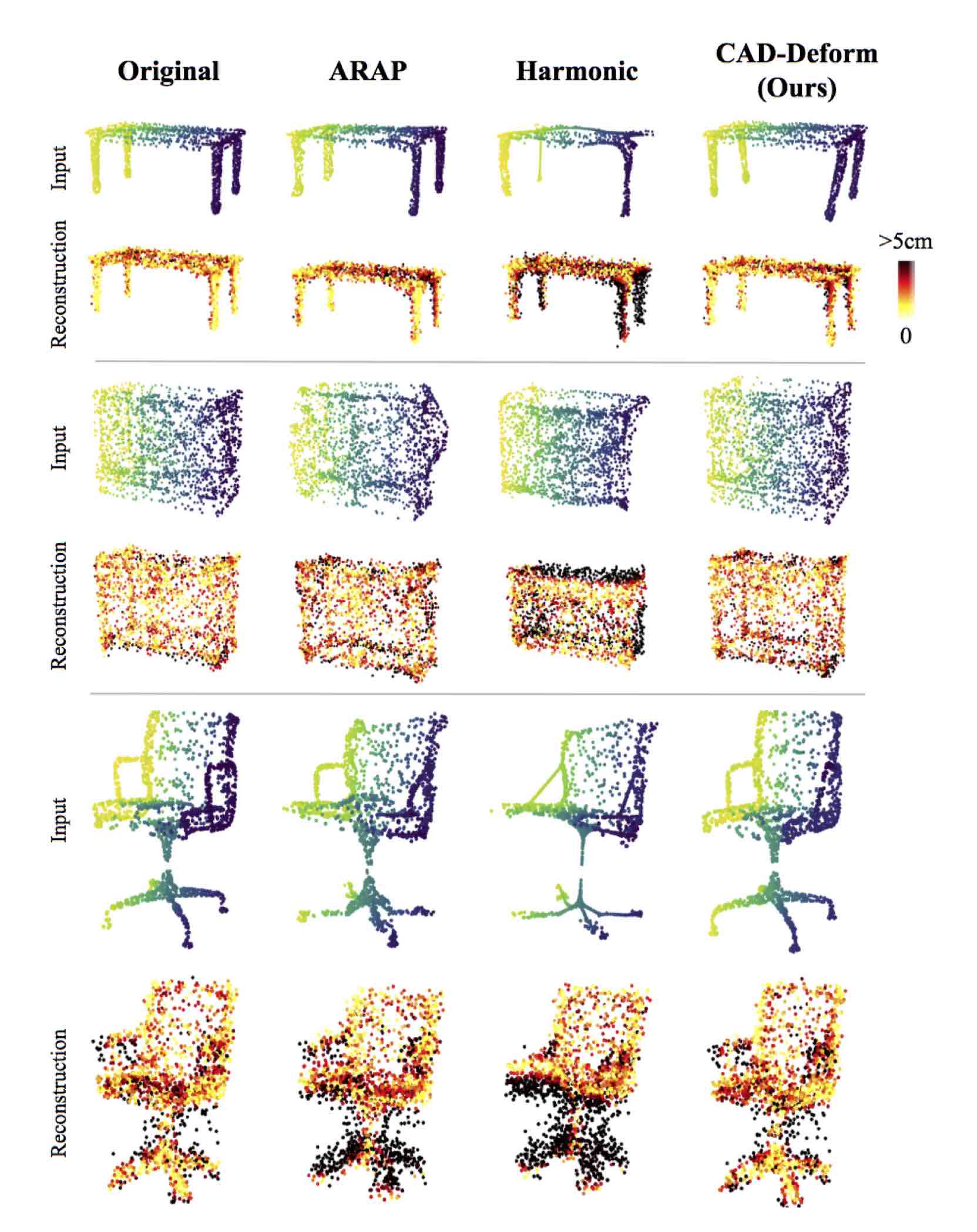}}
\caption{Qualitative comparison of reconstruction of point clouds extracted from mesh vertices. These meshes are obtained using ARAP~\cite{sorkine2007rigid}, Harmonic deformation~\cite{botsch2004intuitive,jacobson2010mixed}, and our CAD-Deform, the first column corresponds to original undeformed meshes. The color of reconstructed point clouds is related to Earth-Mover's Distance between reconstructed and original point clouds of mesh vertices.}
\label{fig:reconstruction_examples}
\end{figure}

\begin{figure}[ht]
\center{\includegraphics[scale=0.24]{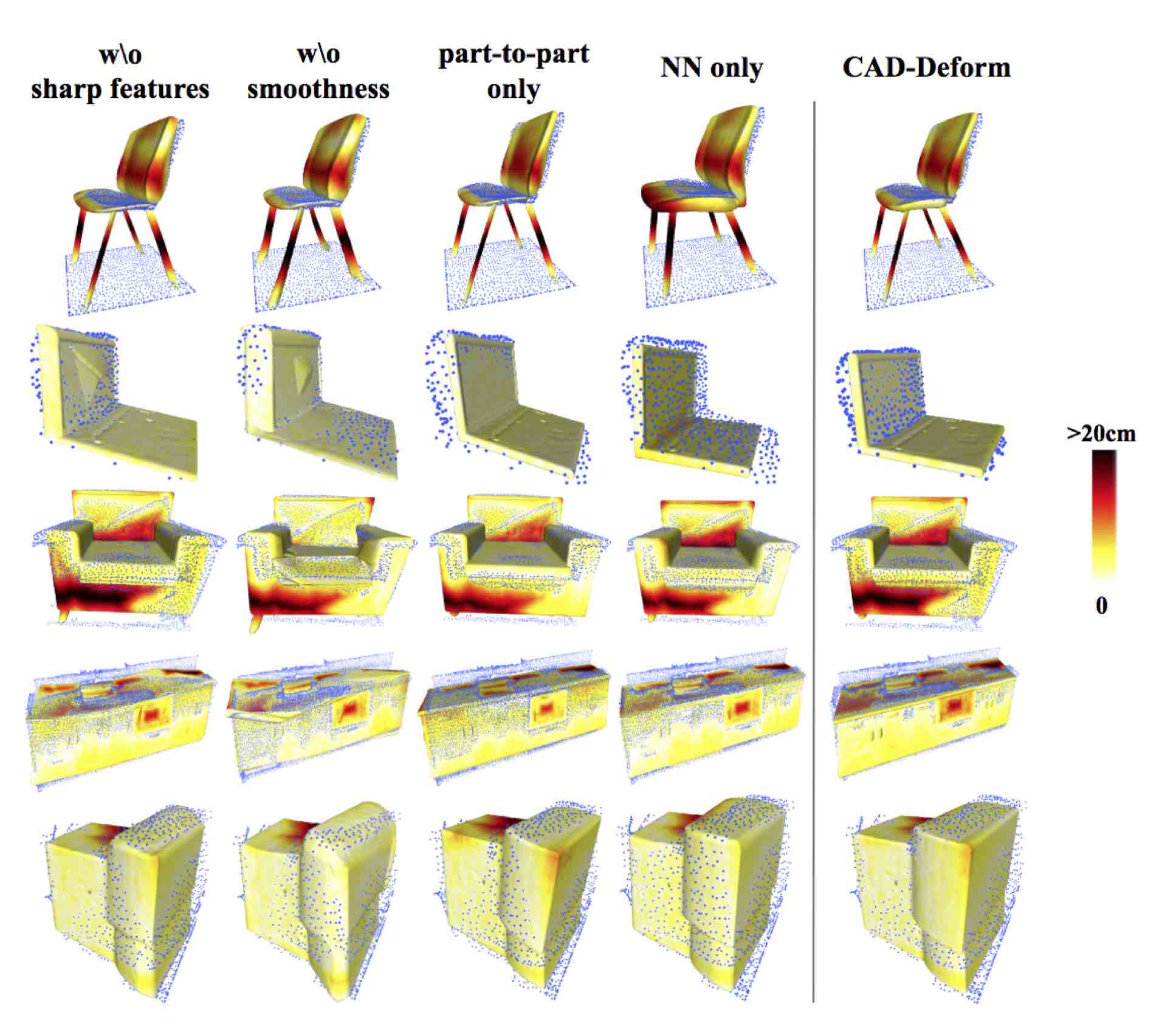}}
\caption{Qualitative results of ablation study usind our deformation framework, with mesh coloured according to the value of the tMMD measure.}
\label{fig:ablation_more_examples}
\end{figure}


In Figure~\ref{fig:fitting_examples}, we display a series of qualitative results with a variety of shape deformations with different classes of instances. Comparing to baselines, our framework achieves accurate fit while preserving sufficient perceptual quality.

Table~\ref{tab:dame_performance} reports the results of surface quality evaluation using deformations obtained with our CAD-Deform vs. the baselines, category-wise. While outperforming the baseline methods across all categories, we discover the smoothness and sharpness energy terms to be the crucial ingredients in keeping high-quality meshes.

Figure~\ref{fig:dame_performance} displays visually the deformation results using the three distinct classes, highlighting differences in surfaces obtained using the three methods.

Table~\ref{tab:reconstruction_performance} reports shape abnormality evaluation results across the six considered categories. Baselines show (Fig.~\ref{fig:reconstruction_examples}) low reconstruction quality as evidenced by a larger number of black points. In other words, comparing to CAD-Deform, the distance from these meshes to undeformed ones is mush larger.

In Figure~\ref{fig:ablation_more_examples}, we show a series of examples for CAD-Deform ablation study. Perceptual quality degrades when excluding every term from the energy.

%% file: tex/suppl_G_morphing.tex
\section{Morphing}

\begin{figure}[ht]
\center{\includegraphics[scale=0.2]{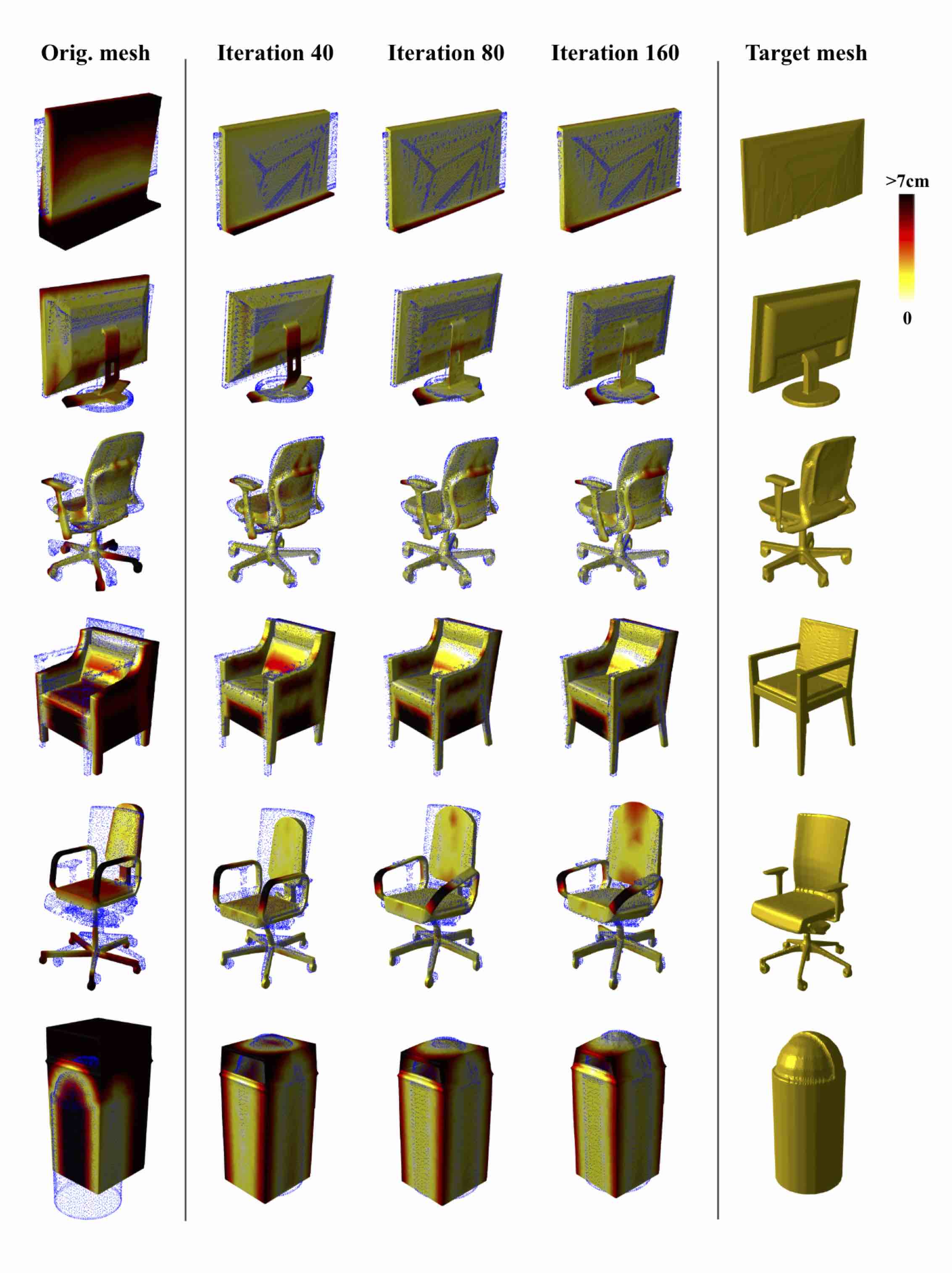}}
\caption{Qualitative shape translation results, interpolating between the original mesh (left) and the target mesh (right).}
\label{fig:morphing_examples}
\end{figure}

In this section, we present an additional series of examples of morphing properties (Fig.~\ref{fig:morphing_examples}). Every iteration of optimization process gradually increases the quality of fit. With CAD-Deform we can morph each part to imitate the structure of the target shape.

%% file: tex/suppl_H_parts.tex
\section{PartNet annotation}

\bgroup
\setlength\tabcolsep{0.5em}
\begin{table}[ht!]
\centering
\resizebox{0.5\textwidth}{!}{%
\begin{tabular}{l cc}
\toprule
\textbf{Accuracy, \%} & class avg. & avg. \\
\midrule
Ground-truth                   & 89.22 & 90.56 \\
Level 1 (object) & 89.25  & 90.79 \\
Level 2               & 89.16 & 91.21 \\
Level 3               & 89.40 & 91.05 \\
Level 4 (parts)  & \bf{91.65} & \bf{93.12} \\
\bottomrule 
\end{tabular}
}
\caption{Comparative evaluation of our approach in terms of Accuracy on different levels of detail.}
\label{tab:part_level_performance}
\end{table}
\egroup

\begin{figure}[ht]
\center{\includegraphics[scale=0.23]{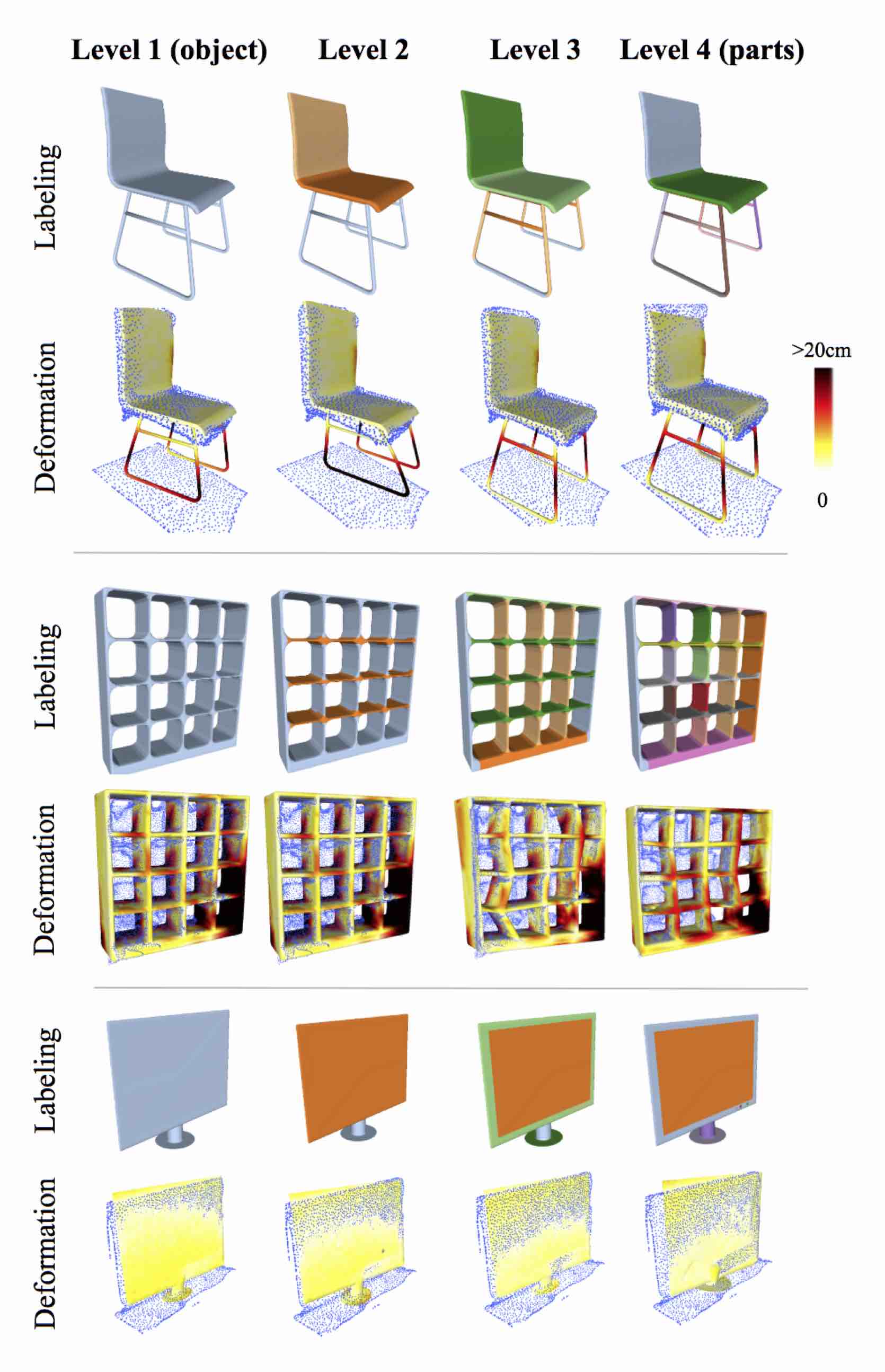}}
\caption{Deformation performance depending on different level of labelling from the PartNet dataset \cite{mo2019partnet}. Deformed mesh surfaces are colored according to the value of tMMD measure, with darker values corresponding to the larger distance values.}
\label{fig:parts_examples}
\end{figure}

This set of experiments shows how quality of fitting depends on mesh vertices labelling. We can provide labels for mesh in diffirent ways depending on the level in PartNet hierarchy \cite{mo2019partnet}. We observe the increase of fitting quality with greater level of detail (Table \ref{tab:part_level_performance}). Examples presented in Figure~\ref{fig:parts_examples} are selected as the most distinguishable deformations on different levels. There are minor visual differences in deformation performance of part labeling level.

%% file: tex/suppl_I_threshold_optimization.tex
\section{Fitting Accuracy analysis}

\begin{figure}[ht]
\center{\includegraphics[scale=0.12]{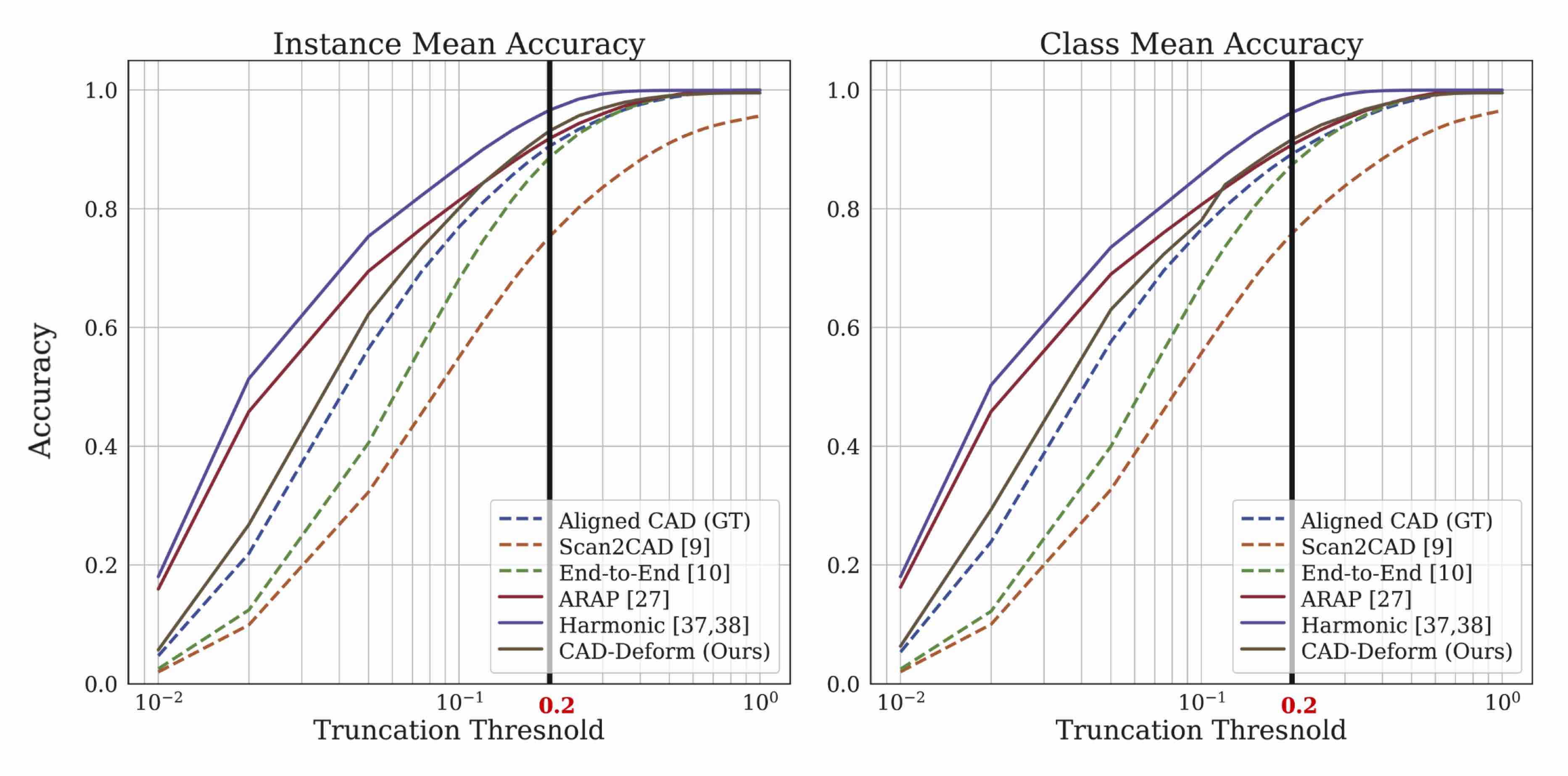}}
\caption{Fitting Accuracy vs. varying $\tau$ threshold for the distance between mesh vertices and close scan points.}
\label{fig:threshold_optimization}
\end{figure}

CAD-Deform deformation framework is sensitive to Accuracy threshold $\tau$ for the distance between mesh vertices and close scan points. In Figure~\ref{fig:threshold_optimization} variation of $\tau$ threshold is presented and we selected $\tau = 0.2\text{~m}$ for fitting Accuracy metric.  

%% file: tex/suppl_J_user_study.tex
\section{Perceptual assessment and user study details}

Having obtained a collection of deformed meshes, we aim to assess their visual quality in comparison to two baseline deformation methods: as-rigid-as-possible (ARAP)~\cite{sorkine2007rigid} and Harmonic deformation~\cite{botsch2004intuitive,jacobson2010mixed}, using a set of perceptual quality measures. The details of our user study design and visual assessment are provided in the supplementary.
To this end, we use original and deformed meshes to compute DAME and reconstruction errors, as outlined in Section~\ref{experiments:setup}, and complement these with visual quality scores obtained with a user study (see below). These scores, presented in Table~\ref{tab:big_picture}, demonstrate that shapes obtained using CAD-Deform have $2\times$ higher surface quality, only slightly deviate from undeformed shapes as viewed by neural autoencoders, and receive $2\times$ higher ratings in human assessment, while sacrificing only 1.1--4.5\,\% accuracy compared to other deformation methods. 

\textit{Design of our user study.} The users were requested to examine renders of shapes from four different categories: the original undeformed shapes as well as shapes deformed using ARAP, Harmonic, and CAD-Deform methods, and give a score to each shape according to the following perceptual aspects: surface quality and smoothness, mesh symmetry, visual similarity to real-world objects, and overall consistency. Ten random shapes from each of the four categories have been rendered from eight different views and scored by 100~unique users on a scale from~1 (bad) to~10 (good). The resulting visual quality scores are computed by averaging over users and shapes in each category.

\begin{figure}[ht]
\center{\includegraphics[scale=0.24]{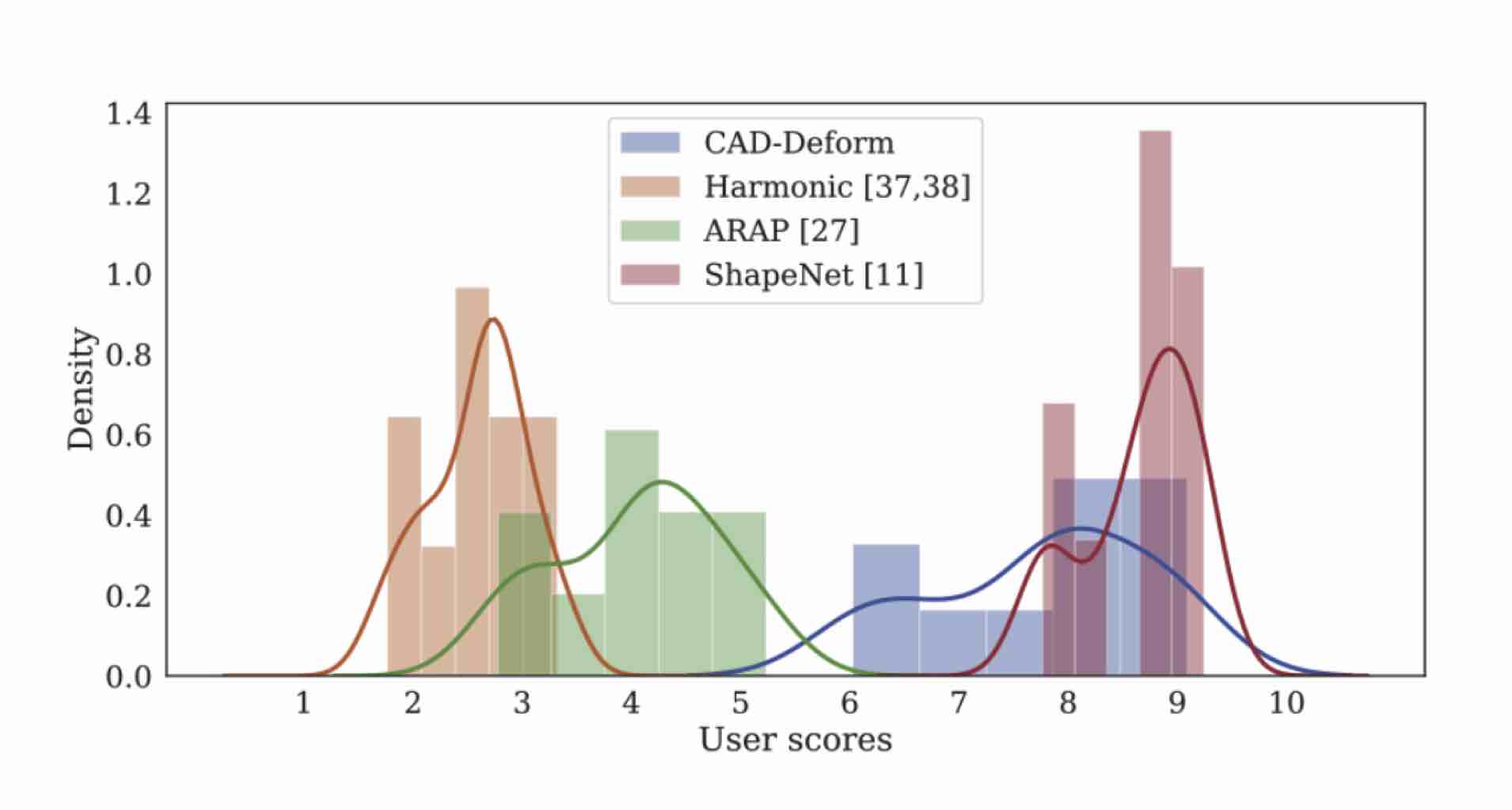}}
\caption{Distribution of user scores averaged by ten shapes from original ShapeNet \cite{chang2015shapenet}, meshes deformed with ARAP \cite{sorkine2007rigid}, Harmonic \cite{botsch2004intuitive,jacobson2010mixed} and CAD-Deform.}
\label{fig:user_study_stats}
\end{figure}

In Figure~\ref{fig:user_study_stats}, we present a distribution of user scores over different deformation methods and shapes. It can be clearly seen that users prefer our deformation results to baselines for all of the cases, which is obvious from the gap between histogram of CAD-Deform and ARAP/Harmonic histograms. At the same time, shapes deformed by CAD-Deform are close to undeformed ShapeNet shapes in terms of surface quality and smoothness, mesh symmetry, visual similarity to real-world objects, and overall consistency. Besides, in Tables~\ref{tab:lap_1}, \ref{tab:lap_2} we provide numbers for evaluation of ARAP/Harmonic deformations w.r.t. the change of Laplacian term weight.

\begin{table}[t]
\centering
\resizebox{0.8\textwidth}{!}{%
\begin{tabular}{l ccc ccc}
\toprule
    \textbf{Lap. term}
    & \multicolumn{3}{c}{\textbf{Class avg.}} 
    & \multicolumn{3}{c}{\textbf{Instance avg.}} \\
    \textbf{weight}
    & \textbf{GT} 
    & \textbf{S2C}~\cite{avetisyan2019scan2cad} 
    & \textbf{E2E}~\cite{avetisyan2019end} 
    & \textbf{GT} 
    & \textbf{S2C}~\cite{avetisyan2019scan2cad} 
    & \textbf{E2E}~\cite{avetisyan2019end} \\

\midrule
$\alpha_{\text{Lap}}=10^{-2}$    & 90.9 & 81.3 & 90.0 & 92.0 & 80.9 & 90.8 \\
$\alpha_{\text{Lap}}=10^{-1}$    & 91.0 & 81.3 & 90.0 & 92.0 & 80.9 & 90.7 \\
$\alpha_{\text{Lap}}=1$         & 91.0 & 81.3 & 89.9 & 91.9 & 80.9 & 90.7  \\
$\alpha_{\text{Lap}}=5$          & 90.9 & 81.2 & 89.9 & 91.9 & 80.9 & 90.7 \\
$\alpha_{\text{Lap}}=20$        & 90.9 & 81.2 & 89.9 & 91.8 & 80.8 & 90.6 \\
\bottomrule
\end{tabular}
}
\caption{Comparative evaluation of ARAP deformations w.r.t. the change of Laplacian term weight in terms of Accuracy (\%).}
\label{tab:lap_1}
\end{table}

\begin{table}[t]
\centering
\resizebox{0.8\textwidth}{!}{%
\begin{tabular}{l ccc ccc}
\toprule
    \textbf{Lap. term}
    & \multicolumn{3}{c}{\textbf{Class avg.}} 
    & \multicolumn{3}{c}{\textbf{Instance avg.}} \\
    \textbf{weight}
    & \textbf{GT} 
    & \textbf{S2C}~\cite{avetisyan2019scan2cad} 
    & \textbf{E2E}~\cite{avetisyan2019end} 
    & \textbf{GT} 
    & \textbf{S2C}~\cite{avetisyan2019scan2cad} 
    & \textbf{E2E}~\cite{avetisyan2019end} \\

\midrule
$\alpha_{\text{Lap}}=10^{-2}$    & 96.3 & 94.3 & 96.6 & 96.7 & 94.5 & 96.9 \\
$\alpha_{\text{Lap}}=10^{-1}$    & 96.3 & 94.2 & 96.6 & 96.7 & 94.4 & 96.9 \\
$\alpha_{\text{Lap}}=1$         & 96.3 & 94.2 & 96.6 & 96.7 & 94.2 & 96.9  \\
$\alpha_{\text{Lap}}=5$          & 96.2 & 94.0 & 96.6 & 96.6 & 94.0 & 96.8 \\
$\alpha_{\text{Lap}}=20$        & 96.2 & 93.8 & 96.5 & 96.6 & 93.8 & 96.7 \\
\bottomrule
\end{tabular}
}
\caption{Comparative evaluation of Harmonic deformations w.r.t. the change of Laplacian term weight in terms of Accuracy (\%).}
\label{tab:lap_2}
\end{table}

